\documentclass[lettersize,journal]{IEEEtran}

\usepackage{amsmath}
\allowdisplaybreaks
\usepackage{amssymb}
\usepackage{mathtools}
\usepackage{amsthm}
\usepackage{enumitem} % itemized

\usepackage{booktabs} % for professional tables

\usepackage{url}
\usepackage{hyperref}

\usepackage[dvipsnames]{xcolor}
\usepackage{algorithm}
\usepackage{algorithmic}

\usepackage{makecell}

\usepackage{caption}
\usepackage{subcaption}

\usepackage{multirow}
\usepackage{multicol}
\usepackage{pifont}% http://ctan.org/pkg/pifont
\newcommand{\cmark}{\color{ForestGreen}\ding{51}}%
\newcommand{\xmark}{\color{red}\ding{55}}%

\usepackage{xr}
\makeatletter
\newcommand*{\addFileDependency}[1]{% argument=file name and extension
\typeout{(#1)}% latexmk will find this if $recorder=0
\@addtofilelist{#1}
\IfFileExists{#1}{}{\typeout{No file #1.}}
}\makeatother

\theoremstyle{plain}
\newtheorem{theorem}{Theorem}[section]

\newtheorem{lemma}[theorem]{Lemma}
\newtheorem{corollary}[theorem]{Corollary}
\theoremstyle{definition}

\newtheorem{assumption}[theorem]{Assumption}
\theoremstyle{remark}
\newtheorem{remark}[theorem]{Remark}

\usepackage[textsize=tiny]{todonotes}

\newcommand{\mL}{ {\mathcal{L} }}

\newcommand{\mG}{ {\mathbf{G} }}

\newcommand{\wm}{{w_m}}
\newcommand{\bw}{{\bf w}}
\newcommand{\tbw}{{\Tilde{\bf w}}}

\newcommand{\mt}{{m_t}}
\newcommand{\tpo}{{t+1}}

\newcommand{\ipo}{{i + 1}}
\newcommand{\BE}[1][{}]{\mathbb{E}_{#1}}
\newcommand{\bu}{\mathbf{u}}
\newcommand{\Var}{\mathrm{Var}}
\newcommand{\hm}{ {h_m} }

\newcommand{\f}[2][{}]{ f_{#1}\left( #2 \right) }

\newcommand{\fmum}[1]{ f_{\mu_m}\left( #1 \right) }

\newcommand{\hn}{ \hat{\nabla} }
\newcommand{\nm}{ \nabla_{w_m}}
\newcommand{\nmt}{ \nabla_{w_\mt}}
\newcommand{\nz}{ \nabla_{w_0}}
\newcommand{\summ}{ {\sum_{m=1}^{M}} }
\newcommand{\sumitau}[1][{}]{ \sum_{i=1}^{\tau {#1}} }
\newcommand{\sumt}{ {\sum_{t=0}^{T-1}} }
\newcommand{\Ls}{L_* }
\newcommand{\ps}{p_* }
\newcommand{\mus}{\mu_* }
\newcommand{\ds}{d_* }
\newcommand{\mGs}{\mG_* }

\newcommand{\bigp}[1]{\left( #1 \right)}
\newcommand{\bigsb}[1]{\left[ #1 \right]}
\newcommand{\bigcb}[1]{\left\{ #1 \right\}}
\newcommand{\NN}{\nonumber \\}
\newcommand{\eq}{ = }
\newcommand{\eqos}[1]{ \overset{ #1}{=} }
\newcommand{\leos}[1]{ \overset{ #1}{\le} }
\newcommand{\norm}[1]{ \left\| #1 \right\|}
\newcommand{\normsq}[1]{ \left\| #1 \right\|^2 }
\newcommand{\ip}[2]{\left< #1, #2\right>}
\newcommand{\uba}[1]{ \underbrace{#1}_{a)} }
\newcommand{\ubb}[1]{ \underbrace{#1}_{b)} }
\newcommand{\ubc}[1]{ \underbrace{#1}_{c)} }
\newcommand{\ubd}[1]{ \underbrace{#1}_{d)} }

\usepackage{algorithmic}
\usepackage{algorithm}
\usepackage{textcomp}
\usepackage{stfloats}
\usepackage{url}
\usepackage{verbatim}
\usepackage{graphicx}

\usepackage{cite}

\begin{document}

\title{Secure and Fast Asynchronous Vertical Federated Learning via Cascaded Hybrid Optimization}

\author{ 
Ganyu Wang, 
Qingsong Zhang,
Li Xiang, %\IEEEmembership{Fellow,~IEEE}, 
Boyu Wang, %\IEEEmembership{Fellow,~IEEE}, \\
Bin Gu, %\IEEEmembership{Fellow,~IEEE},  
Charles Ling, %\IEEEmembership{Fellow,~IEEE},  
        % <-this % stops a space
    \thanks{Bin Gu is with Department of machine learning, Mohamed Bin Zayed University of Artificial Intelligence, Abu Dhabi, UAE (e-mail: jsgubin@gmail.com).}
    \thanks{Charles Ling, Boyu Wang, Xiang Li, Ganyu Wang is with Department of Computer Science of Western University, London, Ontario, Canada. (e-mail: charles.ling@uwo.ca, bwang@csd.uwo.ca, lxiang2@uwo.ca, gwang382@uwo.ca)}
    \thanks{Qingsong Zhang is with School of Electronic Engineering, Xidian University, Xi’an, China (email: qszhang1995@gmail.com).}
    \thanks{Manuscript received June 11, 2023. }
            %revised August 16, 2021.}
}

%\markboth{IEEE Transactions on Neural Networks and Learning Systems, ~Vol.~1, No.~1, June~2023}%

\markboth{Preprint, In Review}%
{Shell \MakeLowercase{\textit{et al.}}: A Sample Article Using IEEEtran.cls for IEEE Journals}

\IEEEpubid{0000--0000/00\$00.00~\copyright~2021 IEEE}
\maketitle

\begin{abstract}

% Background. 
    Vertical Federated Learning (VFL) attracts increasing attention because it empowers multiple parties to jointly train a privacy-preserving model over vertically partitioned data. 
% Central question. 
    Recent research has shown that applying zeroth-order optimization (ZOO) has many advantages in building a practical VFL algorithm. However, a vital problem with the ZOO-based VFL is its slow convergence rate, which limits its application in handling modern large models.
% 1) preserve security via ZOO on Client. 2) speeds up the convergence with FOO on the server. 
    To address this problem, we propose a cascaded hybrid optimization method in VFL. In this method, the downstream models (clients) are trained with ZOO to protect privacy and ensure that no internal information is shared. Meanwhile, the upstream model (server) is updated with first-order optimization (FOO) locally, which significantly improves the convergence rate, making it feasible to train the large models without compromising privacy and security.
% Theoretical result. The ZOO only affect the convergence of the client. 
    We theoretically prove that our VFL framework converges faster than the ZOO-based VFL, as the convergence of our framework is not limited by the size of the server model, making it effective for training large models with the major part on the server.
% Experiment. Practical, if we design a system that has a large model on the server. The convergence can be...
    Extensive experiments demonstrate that our method achieves faster convergence than the ZOO-based VFL framework, while maintaining an equivalent level of privacy protection. Moreover, we show that the convergence of our VFL is comparable to the unsafe FOO-based VFL baseline. Additionally, we demonstrate that our method makes the training of a large model feasible. 

\end{abstract}


\section{Introduction}\label{sec:introduction}

    % introduce FL. HFL. VFL. (引入研究，介绍概念，整体情况。)
    Data availability is essential for machine learning, however, privacy concerns often prevent the direct sharing of data among different parties. Federated learning (FL) addresses this issue by facilitating collaborative model training without sharing private data. 
    This approach allows multiple parties to leverage their data while adhering to the privacy protection measure and the government regulation, such as the General Data Protection Regulation (GDPR)~\cite{EUdataregulations2018}.

    % 介绍 HFL and VFL. 
    FL algorithms have evolved into two mainstream sub-types, Horizontal Federated Learning (HFL)~\cite{mcmahan2017communication, li2020federated, li2021fedbn, karimireddy2020scaffold, mishchenko2022proxskip} and Vertical Federated Learning (VFL)~\cite{li2020federatedservey, vepakomma2018split, chen2020vafl, yang2019quasi, hu2019fdml, wei2022vertical, gu2021privacy}. HFL involves clients holding a subset of data points with a full feature set (horizontally distributed), while VFL involves clients holding all data points but with a non-intersecting subset of features (vertically distributed).

    % 深入介绍VFL。
    We focuses on VFL, which is applicable to practical learning scenarios in various industries, such as hospitals, banks, and insurance companies. For example, a government agency (server) collaborates with multiple banks (clients) to develop a model for estimating customers' credit scores~\cite{wei2022vertical}, where each bank holds a distinct set of customer features. 
    In VFL, the client trains a feature extraction model that maps its local data sample to embeddings. The server then collects the embeddings from all clients and uses them as input for the server model to make a prediction.

\begin{figure*}[t]
    \centerline{\includegraphics[width=\linewidth]{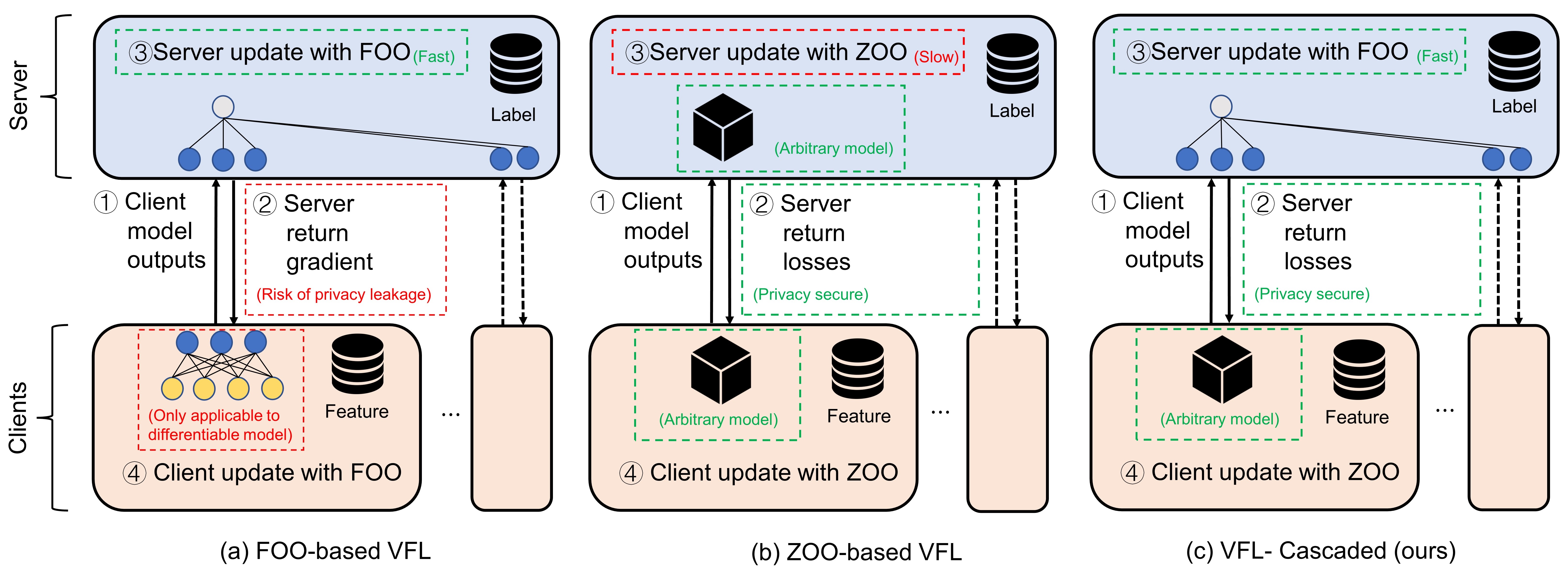}}
    \caption{The Intuition of our VFL Framework}
    \label{fig:framework_general}
    \end{figure*}

    % The four metrics. 
    To build a practical VFL framework, it is essential to meet the following fundamental requirements: model applicability~\cite{castiglia2022flexible, makhija2022architecture, zhang2021desirable}, privacy security~\cite{zhou2020vertically, hardy2017private, fang2021large}, computational efficiency~\cite{chen2020vafl, hu2019fdml, zhang2021desirable}, and communication efficiency~\cite{zhang2021desirable, castiglia2022compressed, wang2022communication}. 
    % We combine the best among two different optimization methods. FOO and ZOO
    % 图片讨论，两种不同的optimization VFL 的优缺点。
    In a typical VFL framework optimized with FOO~\cite{chen2020vafl, vepakomma2018split}, as illustrated in Figure~\ref{fig:framework_general} (a), both the server and clients utilize FOO to optimize the model, which is fast. However, sharing the gradient with the client poses a serious risk of privacy leakage~\cite{fu2022label, fredrikson2015model, he2016deep, zhao2020idlg}, and the framework is only applicable to differentiable models.

    %--------------------------------------------------------
    \IEEEpubidadjcol % Change the height of the second column so that the "pubid" does not overlap with the body text.
    %--------------------------------------------------------

    A recent study~\cite{zhang2021desirable} found that applying ZOO on VFL, as depicted in Figure~\ref{fig:framework_general} (b), offers several advantages in building practical VFL. Firstly, it enhances model applicability by eliminating the requirement for an explicit gradient to update the model. 
    Secondly, it improves privacy security by transmitting black-box information (losses) with the client instead of internal information (gradients).
    Besides, the client retains the perturbation direction, preventing third parties from obtaining the gradient. As a result, both the server and client can maintain the confidentiality of gradient information during training. However, relying solely on ZOO for model optimization can lead to slow convergence, especially when dealing with large models.
    
    % Summarize the problem. 
    Both frameworks mentioned above do not meet the requirement of practical VFL. 
    Although FOO converges rapidly and dependably, the privacy risk associated with transmitting the gradient is a significant drawback. 
    On the other hand, ZOO provides high model applicability and privacy security, but suffers from slow convergence problem.
    % raise the question. Gap. 
    Then, it comes to the question: \textit{How to improve the convergence speed while preserving the advantages of ZOO to make a practical VFL?}
    
    % We propose our framework by combining the best of the two optimization methods.  
    In this paper, we provide a solution to this problem by proposing a cascaded hybrid optimization method in the asynchronous VFL which maximizes the benefits of both optimization methods. 
    % Our method in Fig. 1 (c). 
    As depicted in Figure~\ref{fig:framework_general} (c), we utilized distinct optimization methods for the upstream (server) and downstream (client) of the global model in a cascaded manner. 
    This approach ensures privacy preservation, as the downstream models update with ZOO, which guarantees that no gradient is transmitted through the network. Additionally, the upstream model is updated with FOO locally, which converges fast and does not compromise privacy. 

    % Summarize the contribution. 
    Our contributions can be summarized as follows: 
    \begin{itemize}[topsep=0pt] 
        \setlength\itemsep{-0.1em}
        \item We propose a practical asynchronous VFL framework that cascades two different optimization methods (FOO\&ZOO), where the advantages of both optimization methods were maximized. Our VFL framework satisfy the fundamental requirements of model applicability, privacy security, computational efficiency, and communication efficiency to a significant degree.
        \item We theoretically prove that the convergence of our VFL framework is faster than the ZOO-based VFL by demonstrating that the convergence is solely limited by the size of the client's parameters. Additionally, our VFL framework can feasibly train a large parameterized model with the majority of the model located on the server.
        \item We conduct extensive experiments on the multi-layer perceptron model, Convolutional Neural Network, and transformer model (BERT) to demonstrate the privacy and applicability of our framework in the latest deep learning tasks.
    \end{itemize}

    % The rationale for the setting.
    \textbf{Justification of the Application Scenario:}
    In our VFL setting, the server uses a larger model compared with the clients, we provide our justification for this application scenario below.
    
    In VFL, the server is typically the initiator and primary beneficiary of the model training process. 
    The client, on the other hand, acts as a follower and only provides the embedding of their local features without disclosing the raw data~\cite{wei2022vertical}. 
    Besides, the server usually possesses more computational resources than the clients, making it more suitable for training large models. 
    Therefore, using a larger model on the server side can lead to better data predictions and reduce the computational burden for all participants in the VFL, making it a more preferable and economical option.
    
    % Since only the server can get the prediction result, it is reasonable that the server uses a larger model, while the client holds a smaller model. 

\section{Related Work}
    % (Review related work and identify the shortcomings of existing work. 1. General shortcomings. 2. Shortcomings of each individual work.)
    %  Related works about 4 metrics: 
    %  model applicability,  privacy security, computation efficiency, and communication efficiency.
    There are several basic metrics to consider when developing and evaluating a VFL framework:
    
    \textbf{Model Applicability} dictates the VFL framework can fit heterogeneous models. The heterogeneity of the model mainly determines whether the model is differentiable. 
    For example, most of the VFL approaches explicitly apply gradient~\cite{vepakomma2018split, chen2020vafl}, which enforce each party to use a differentiable model. 
    However, this approach may not always be practical, especially when the participants have non-differentiable model architectures. 
    In such cases, when the gradient is not available, the main solution is to apply proximal-term~\cite{castiglia2022flexible} or to use zeroth-order optimization~\cite{zhang2021desirable}.

    \textbf{Privacy} is a critical consideration for any VFL algorithm. In VFL, there are two types of private data: the features held by the clients and the labels held by the server. Depending on the target of the attack, privacy inference attacks in VFL can be classified as feature inference attacks~\cite{luo2021feature, jin2021cafe, zhu2019deep, fredrikson2015model, weng2020privacy} or label inference attacks~\cite{fu2022label, sun2022label, zhu2019deep, zhao2020idlg, jin2021cafe}.
    
    The mainstream privacy protection scheme is applying privacy computing on VFL. For example, Liu et al.~\cite{liu2020boosting} and Hardy el al.~\cite{hardy2017private} have applied homomorphic encryption (HE) on the transmission data, where the participant in the VFL framework sends the ciphertext instead of plain text through the network. 
    Other works have used differential privacy (DP)~\cite{shokri2015privacy, ranbaduge2022differentially, wei2020federated} or secure multiparty computation (SMC)~\cite{fang2021large}. 
    Although these privacy computing methods have a provably security level, they have several disadvantages. 
    For example, HE restricts the choice of model structure, DP reduces the performance of the global model, and HE and SMC have high communication or computation costs for participants, which limits their application.

    \textbf{Computational Efficiency} dictates that the computation resource in VFL is efficiently used. The computational efficiency of synchronous VFL can be low due to the idle time for participants. 
    In synchronous VFL, the server coordinates with all clients by sending a request to all clients for each batch of training data.
    The server must wait for all clients' responses to fulfill one global update step before sending the next request to all clients~\cite{liu2019communication, vepakomma2018split, castiglia2022compressed, fang2021large}.
    As a result, all participants must wait for the slowest one, leading to low computational efficiency in synchronous VFL.
    
    Asynchronous VFL~\cite{chen2020vafl, hu2019fdml, zhang2021desirable} was proposed to reduce idle time for each participant and improve the computation efficiency. In asynchronous VFL, the client continuously sends its model output to the server without coordination from the server. When the server receives the output from the client, it replies with the necessary information (e.g., partial derivative) to assist the model update of the client. This scheme eliminates most of the idle time for the clients and improves computation efficiency. Our research focuses on asynchronous VFL. 
    % However, due to asynchronous client updates, the convergence rate of the global model is limited by the delay of the slowest clients, which may reduce the training efficiency. 

    \textbf{Communication Efficiency} is about reducing the communication cost between the parties of VFL. 
    Research has focused on reducing communication rounds~\cite{liu2019communication, wang2022communication} or per-round communication overhead~\cite{castiglia2022compressed}. 
    Liu et al.\cite{liu2019communication} propose multiple local updates on VFL participants to reduce communication rounds. However, multiple local updates consume more computational resources on clients, which is not favorable in VFL. 
    Wang et al.\cite{wang2022communication} apply a better optimization method to speed up convergence and reduce communication rounds. 
    Castiglia et al.~\cite{castiglia2022compressed} apply compression to the embeddings of client outputs to support efficient communication and multiple local updates, reducing per-round communication overhead and communication rounds.
    % Castiglia et al. \cite{castiglia2022compressed} use compression on the embeddings of the client outputs to support efficient communication.  

\section{Method} \label{sec:method}
    This section introduces the modeling of the VFL problem and proposes our VFL framework that cascaded different optimization methods. With a cascaded hybrid optimization method, the advantage of both ZOO and FOO is maximized in one VFL framework. 

    \subsection{Problem Definition}
        % intro Server and client. 
        We consider a general form of VFL problem ~\cite{chen2020vafl, hu2019fdml, liu2019communication, zhang2021desirable}, which involves a single server and $M$ clients. 
        % Dataset. The intro client holds the data. Intro Server holds the label. 
        Each participant in the VFL possesses $n$ samples within their respective databases. Specifically, each client holds a distinct set of features for each sample, denoted as $x_{i,m}$, while the server holds the corresponding labels for the $i$-th sample\footnote{For brevity, we use a single data sample $i$ for discussion, however, the discussion can be easily generalized to a mini-batch version.}, denoted as $y_i$. 

        % Data not sharing. client model. Server model. Loss function.
        Clients communicate with the server through the network. To preserve the privacy of the local data. Raw data $x_{i, m}$ and $y_i$ should not be transmitted through the network. The client holds a local model $ F_m( w_m;x_{i, m}) $ parameterized by $w_m \in \mathbb{R}^{d_m}$ with sample $x_{i, m}$ as input and send the output $c_{i, m}$ of the model to the server through the network. The server holds a model $F_0(w_0; c_{i, 1}, \hdots, c_{i, q})$ which is parameterized by $w_0 \in \mathbb{R}^{d_0} $ and take $c_{i, m}$ from all clients as inputs. The loss function is denoted as $\mathcal{L}( \hat{y_i}, y_i)$. 
        
        % the form of the problem. 
        Ideally, all parties in the VFL framework collaborate to solve a finite-sum problem in the composition form:
        % Formula
        \begin{align} \label{eq:problem}
              f(w_0, \bw) 
            =& \frac{1}{n} \sum_{i=1}^{n} \underbrace{\bigsb{ \mL (F_0(w_0, c_{i, 1}, \hdots, c_{i, M}), y_i) + \lambda \sum_{m=0}^M g(w_m)} }_{f_i(w_0, \bw) } \NN
             & \text{with \quad } c_{i, m} = F_m ( w_m ;x_{i, m})  \quad \forall m \in [M] 
        \end{align}
        where $g$ is the regularization function for the party $m$, $[M]$ $=$ $\{1, 2, \cdots, M\}$ denote the set of all clients' indices, $\bw$ $=$ $\{ w_1, w_2, \cdots, w_M \}$ denotes the parameters from all clients, $f_i(w_0, \bw)$ denotes the loss function for the $i$-th sample. 
        
        % Application of the "general form of the problem" in Machine Learning. 
    \subsection{Cascaded Hybrid Optimization (ZOO\&FOO)}
        % The privacy of applying ZOO in VFL. 
        To leverage the advantage of ZOO and FOO in one VFL, we apply a cascaded hybrid optimization method, where the upstream (server) and the downstream (client) of the global model applies different optimization method simultaneously. Specifically, the clients are updated with ZOO and the communication between the server and the client does not contain internal information, which protects privacy security. The server is updated with FOO locally, which speeds up the convergence of the VFL without degrading the privacy security.  
        % The client interacts with the server to update its model, where the client sends the model outputs (embeddings) to the server, and the server returns the losses w.r.t. the client's embeddings.
        
        \subsubsection{Client Update with ZOO to Ensure Privacy Security}
            % General. two-point estimator definition. 
            The models of the clients are trained with the zeroth-order optimization. The two-point stochastic gradient estimator \cite{liu2020primer, nesterov2017random} w.r.t. the client $m$'s parameter $w_m$ is defined as:
            % two-point estimator Eq. 
            \begin{align} \label{eq:gradient_estimator_ori}
                  &\hn_{w_m} f_i \bigp{w_0, \bw} 
                = \frac{\phi(d_m) }{\mu_m} \bigsb{  f_i\bigp{w_m + \mu_m u_{i, m}}  - f_i\bigp{w_m}} u_{i, m} 
            \end{align}
            where $u_{i, m} \sim p$ is a random direction vector drawn from distribution $p$. Typically, $p$ is standard normal distribution $\mathcal{N}(\pmb{0}, \mathbf{I})$, or uniform distribution $\mathcal{U}(\mathcal{S}(0, 1) )$ over a unit sphere at $\mathbf{0}$, with the radius of $1$. $\mu$ is the smoothing parameter. $f_i\bigp{w_m + \mu_m u_{i, m}}$ is the simplified form of $f_i(w_0, w_1, w_2, \cdots, w_m + \mu_m u_{i, m}, \cdots, w_q )$ , i.e. the loss of the $i$-th sample with the model parameter of client $m$ changed to $w_m + \mu_m u_{i, m}$. $\phi(d_m)$ is a dimension-dependent factor that relates to the choice of $p$. To be more specific, if $p$ is $\mathcal{N}(\pmb{0}, \mathbf{I})$ then $\phi(d_m) = 1$ and if $p$ is $\mathcal{U}( \mathcal{S}(0, 1))$ then $\phi(d_m) = d_m$.
    
            % how to get the two-point gradient estimator. Process 
            The clients are unable to compute the gradient of the loss function locally due to the fact that the label of the data is stored on the server. As illustrated in Figure~\ref{fig:framework_FO}, the clients query the server for the necessary computation material. 
            The active client then computes the model output with or without the perturbation $\mu_m u_{i,m}$ on its parameter and sends them to the server. Specifically, the client's outputs are:
            \begin{align}
                c_{i, m}  & = F_m (w_m; x_{i,m}) \NN
                \hat{c}_{i, m} & = F_m(w_m + \mu_m u_{i, m} ; x_{i,m})  \nonumber
            \end{align}
            
            Receiving the query from the client, the server replies to the client $m$ with the corresponding loss values $h_{i, m}$ and $\hat{h}_{i, m}$:
            \begin{align}
                h_{i, m}       & = \mL (F_0(w_0, c_{i, 1}, \hdots, c_{i, m}, \hdots, c_{i, M}), y_i) \NN
                \hat{h}_{i, m} & = \mL (F_0(w_0, c_{i, 1}, \hdots, \hat{c}_{i, m}, \hdots, c_{i, M}), y_i) \nonumber
            \end{align}
            
            When the client receives $h_{i, m}$ and $\hat{h}_{i, m}$ from the server, it is able to calculate the two-point gradient estimator via:
            \begin{align} \label{eq:gradient_estimator}
                  \hn_{w_m} f_i \bigp{w_0, \bw} = \frac{\phi(d_m) }{\mu_m} \bigsb{ \hat{h}_{i, m} - h_{i, m} } u_{i, m} 
            \end{align}
            % gradient descent. 
            Finally, the client $m$ updates its parameter by gradient descent with the stochastic gradient estimator:
            \begin{align}
                w_m^\tpo = w_m^t - \eta_m \hn_{w_m} f_i \bigp{w_0^t, \bw^t} \nonumber
            \end{align}

            There are two parts of private data in the VFL framework that require protection: the features held by the clients and the labels held by the server.
            Our framework protects the privacy of the data by concealing the internal information of the participants. 
            A comprehensive analysis of the privacy protection of our framework is presented in section~\ref{sec:security_analysis}.

        \subsubsection{Server Update with FOO to Solve the Slow Convergence Problem}
            % General introduction. slow convergence problem. 
            The primary issue with ZOO in the context of recent data mining tasks is that the variance of the gradient estimation increases as the parameter dimension grows larger, leading to slow convergence of ZOO, particularly for large models.
            To address this issue, we implement the FOO on the server to speed up the convergence. 
            It is important to note that the server update is performed locally and does not affect communication with the client or the client's update steps. 
            As a result, the privacy protection of the framework is not compromised while simultaneously accelerating convergence.

            % General. Update timing. Explicit. 
            The server's model is trained with the first-order gradient. Whenever the server receives a message from the client, it performs one gradient descent step on its local model. Since the server can access the output embeddings $[c_{i, m}]_{m=1}^M$ from all clients and the label $y_i$, plus that the server naturally has full access to its own model $F_0$, the server can explicitly calculate the gradient via backpropagation. Specifically, the local gradient of the server is: 
            \begin{align} % \label{eq:server_gradient}
                \nz f_{i} (w_0, \bw) = \frac{\partial \bigsb{ \mL (F_0(w_0, c_{i, 1}, \hdots, c_{i, M}), y_i) + \lambda g(w_0)} }{\partial w_0} \nonumber
            \end{align}
            And the server's parameter is updated via gradient descent:
            \begin{align}\label{eq:server_update}
                w_0^\tpo = w_0^t - \eta_0 \nz f_{i} (w_0^t, \bw^t) 
            \end{align}

    \subsection{Asynchronous Updates}

\begin{figure}[tbp]
            \centerline{\includegraphics[width=0.9\linewidth]{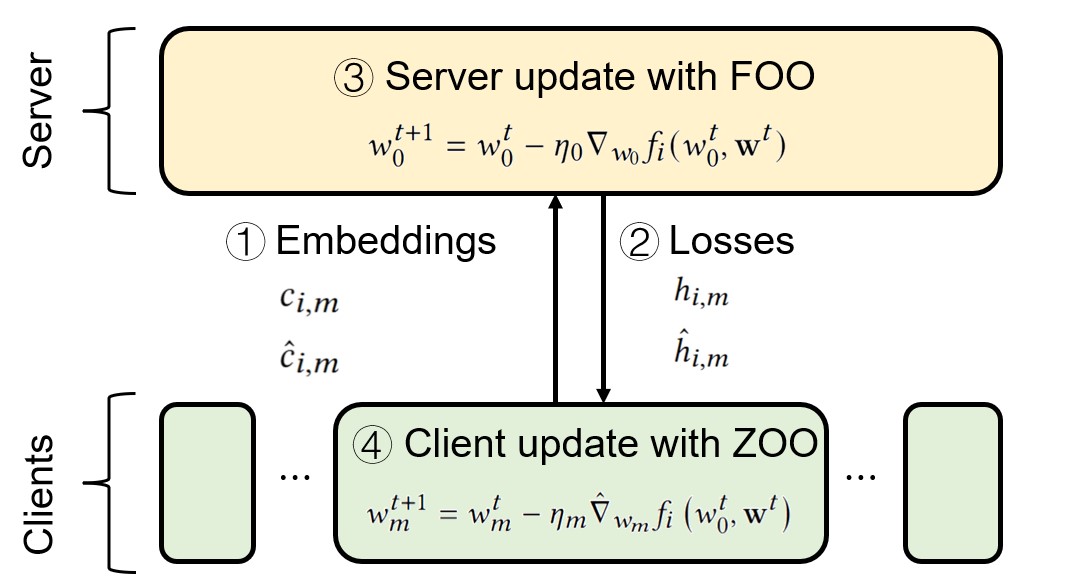}}
            \caption{One Round of our VFL Framework}
            \label{fig:framework_FO}
            \vspace{1pt}
        \end{figure}

        % No coordination. Assumption of a reliable network (no message miss). Modeling with update sequence. 
        The global model is trained without coordination among each party. 
        We assume that all messages will be successfully transmitted, and no participants will withdraw during training. 
        A schematic graph is shown in Figure~\ref{fig:framework_FO}. At each round, only one client is activated and communicates with the server. After the communication, the activated client and the server update their model. The clients' update order can be modeled with a sequence of length $T$. In the $t$-th iteration, the client $m_t$ is activated and picks the $i$-th sample for the update.  

        % modeling the delay. 
        To model the delay of the clients, if the client $m_t$ is activated at the $t$-th iteration, the client updates its parameter $w_\mt$ and its delay for the $i$-th sample on the global model is reset. For all other clients $m \neq \mt $, the delay count is incremented by $1$. Formally, the delay for the client $m$ and sample $i$ is updated using the following equation:
        
        \begin{align}
            \tau^{t+1}_{i, m} = \begin{cases}
                1, & \quad m = m_t, i = i_t\\
                \tau^{t}_{i, m} + 1, &\quad \text{otherwise} \nonumber
            \end{cases}
        \end{align}
        Taking the client delay $\tau^{t}_{i, m}$ into consideration, we can represent the set of parameters for the delayed clients as:
        \begin{align}
            \tbw^t = {{{\bf w}}}^{t-\tau^{t}_{i}} = [w_1^{t- \tau^t_{i, 1} }, \hdots, w_M^{t-\tau^t_{i, M}} ] \nonumber
        \end{align}
        
        % The synchronized case. 
        % For example, in Vepakomma et al.'s~\cite{vepakomma2018split} work, the server coordinates with each client. In this case, the VFL problem can be seen as a model learning in regular training procedures but with the separated parts of the model holder in each client. 

        %\begin{figure}[htbp]
        %    \centering
        %    \includegraphics[width=0.9\linewidth]{Figure/Framework_FO.jpg}
        %    \caption{One round of our VFL framework}
        %    \label{fig:framework_FO}
        %\end{figure}

    \subsection{Algorithm}
        \begin{algorithm}[htbp]
    	\caption{Asyn. VFL with Cascaded Hybrid Optimization}\label{algo:1}
    	\begin{algorithmic}[1]
    		\item Initialize variables for workers $m \in [M]$
    		\WHILE{ not convergent}
    		\STATE \textbf{when} a \textbf{client} $m$ is activated, {\bf do}:
    		\STATE \ \ Randomly select a sample $x_{i, m} $
    		\STATE \ \ Compute $c_{i,m}$, $\hat{c}_{i,m}$ and  upload them to the server
    		\STATE \ \  Receive $h_{i,m}$ and  $\hat{h}_{i,m}$ from the server (in a listen manner)
    		\STATE \ \  Compute $ \hn_{w_m} f_i(w_0, \tbw) $ via Eq.~\ref{eq:gradient_estimator} (ZOO)
    		\STATE \ \ Update $w_m\leftarrow w_m-\eta_m \hat{v}_m $
    		\STATE {\bf when server} receives $c_{i,m}$ and $\hat{c}_{i,m}$, {\bf do}:
    		\STATE \ \ Compute and send $h_{i,m}$, $\hat{h}_{i,m}$ to client $m$
    		\STATE \ \  Compute $ \nabla_{w_0} f_i(w_0, \tbw) $,~ (FOO)
    		\STATE \ \ Update $w_0\leftarrow w_0-\eta_0 \nabla_{w_0} f_i(w_0, \tbw)$
    		\ENDWHILE
    	\end{algorithmic}
        \end{algorithm}
    
        % The framework. Fig, framework. 1 2 3 4 steps. 
        By combining the ZOO on the client and FOO on the server, we designed an asynchronous VFL framework. 
        The algorithm is presented in Algorithm~\ref{algo:1}, and the procedure of one update round is summarized in Fig.~\ref{fig:framework_FO}. 
        The procedure of each training round can be summarized as follows: first, the client randomly selects one sample $i$, computes $c_{i, m}$ and $\hat{c}_{i, m}$, and sends them to the server. 
        Upon receiving the query from client $m$, the server calculates the corresponding losses $h{i, m}$ and $\hat{h}_{i, m}$ and sends them back to the client. 
        The server updates its parameter using gradient descent (Eq.\ref{eq:server_update}) immediately after sending the losses to the client. 
        Finally, upon receiving $h_{i, m}$ and $\hat{h}_{i, m}$ from the server, the client updates its parameter using the stochastic gradient estimator given by Eq.~\ref{eq:gradient_estimator}.
        % after which the server updates its parameter with FO. (for adjusting length.)

\section{Convergence Analysis} \label{sec:convergence}
    \subsection{Theoretical Difficulty}
        The theoretical difficulty of our work comes from the cascaded hybrid optimization in the VFL, where different optimization methods are simultaneously applied to the upstream and downstream parts of the VFL. 
        To the best of our knowledge, all related works in VFL have only considered a single type of optimization method in the entire VFL during one iteration, whose analytic result can be more easily derived via the same analytic steps on the entire framework. 
        However, our work required different analytic procedures to be applied to different parts of the model to solve the problem, which posed a significant challenge. 
        Specifically, the analytic procedure for ZOO and FOO is vastly different, making it difficult to analyze these two different optimizations in one model.

    \subsection{Assumptions}
        % Assumptions. 
        Assumptions \ref{assum:feasible_optimal_solution} - \ref{assum:bounded_variance} are the basic assumptions for solving the non-convex optimization problem with stochastic gradient descent \cite{ghadimi2013stochastic, liu2019communication, zhang2021desirable}. Assumption~\ref{assum:feasible_optimal_solution} tells that the global minima $f^*$ is not $-\infty$~\cite{ghadimi2013stochastic, liu2018zeroth, zhang2021desirable}. Assumption~\ref{assum:smoothness} is used for modeling the smoothness of the loss function $f(\cdot)$, with which we can link the difference of the gradients with the difference of the input in the definition domain. Assumption~\ref{assum:unbiased_gradient} is a common assumption for stochastic gradient descent telling that the expectation of the estimation of the stochastic gradient of the sample $i$ does not have a systematic error or bias~\cite{ghadimi2013stochastic}. Assumption~\ref{assum:bounded_variance} tells that the variance of the gradient estimation is bounded~\cite{liu2018zeroth}. 
        
        \begin{assumption}\label{assum:feasible_optimal_solution}
        {\bf{Feasible Optimal Solution:}} Function $f$ is bounded below that is, there exist $f^*$ such that,
        \begin{align}
            f^*:=\inf_{[w_0,{{\bw }}]\in \mathbb{R}^d} f(w_0,{{\bw}}) > -\infty. \nonumber
        \end{align}
        \end{assumption}
        \begin{assumption}\label{assum:smoothness}
             {\textbf{Lipschitz Gradient:}} $\nabla f_i$ is $L$-Lipschitz continuous w.r.t. all the parameter, i.e., there exists a constant $L$ for $ \forall \ [w_0,{{\bf w}}], [w_0', {{\bf w}}']$  such that
            \begin{align}
                & \norm{ \nabla_{[w_0,{{\bf w}}]} f_i (w_0, \bw) - \nabla_{[w_0,{{\bf w}}]} f_i (w'_0, \bw')} \NN 
                & \qquad \qquad\qquad\qquad \le L \norm{[w_0,{{\bf w}}]-[w_0',{{\bf w}}']} \nonumber
            \end{align}
            specifically there exists an $L_m>0$ for all parties $m=0,\cdots,M$ such that $\nm f_i$ is $L_m$-Lipschitz continuous:
            \begin{align} 
                    & \norm{ \nm f_i (w_0,\bw) - \nm f_i (w_0',\bw')} \NN 
                    & \qquad \qquad\qquad\qquad \le L_m \norm{[w_0,{{\bf w}}]-[w_0',{{\bf w}}']} \nonumber
            \end{align}
        \end{assumption}
        
        \begin{assumption}\label{assum:unbiased_gradient}
            {\textbf{Unbiased Gradient:}} For $m \in 0, 1, \cdots M $ for every data sample $i$, the stochastic partial derivatives for all participants are unbiased, i.e.  
                $$ \BE[i] \nm f_i(w_0, \bw ) = \nm \f{w_0, \bw} $$
        \end{assumption}
    
        \begin{assumption}\label{assum:bounded_variance}
            \textbf{Bounded Variance:} For  $ m = 0, 1, \cdots, M $, there exist constants $\sigma_m \le \infty $ such that the variance of the stochastic partial derivatives are bounded: 
                $$\BE[i] \norm{\nm f_i(w_0, \bw) - \nm f(w_0, \bw) }^2 \leq \sigma_m^2 $$
        \end{assumption}
    
        Assumption~\ref{assum:bounded_embedding_gradients} is a common assumption for analyzing VFL when bounding some terms for the entire model when the rest parts have been bounded \cite{castiglia2022compressed, gu2021privacy, zhang2021secure}. We only apply this assumption in the parts of convergence analysis that do not affect the analytic result. 
        \begin{assumption}\label{assum:bounded_embedding_gradients}
            \textbf{Bounded Block-coordinate Gradient:} The gradient w.r.t. all the client is bounded, i.e. there exist positive constants $\mathbf{G}_m$ for the client $ m = 1, \cdots, M $ the following inequalities hold: 
            \begin{align}
                \norm{ \nm \hm(\wm; x_{m, i}) } \le \mG_m  \nonumber
            \end{align}
        \end{assumption}
    
        % Asynchronous Update. 
        Assumption~\ref{assum:independent_client}-\ref{assum:bound_delay} are fundamental and widely used assumptions for analyzing the asynchronous VFL~\cite{zhang2021desirable, chen2020vafl, gu2021privacy}. 
        Asynchronous updates introduce delays for clients when calculating losses. 
        We assume that the activation of clients at each global round is independent (assumption~\ref{assum:independent_client}) and that the maximum delay is bounded (assumption~\ref{assum:bound_delay}), which are reasonable assumptions for modeling asynchronous VFL. 
        Without a bounded delay assumption, convergence cannot be achieved, and the convergence result cannot be further simplified because the distribution of client activation is unknown.
        
        \begin{assumption}\label{assum:independent_client} {\bf{Independent Client:}}
            The activated client $m_t$ for the global iteration $t$ is independent of $m_0$, $\cdots,$ $m_{t-1}$ and satisfies $\mathbb{P}(m_t=m):=p_m$
        \end{assumption}
        
        \begin{assumption}\label{assum:bound_delay}{\bf{Uniformly Bounded Delay:}} For each client $m$, and each sample $i$, the delay at each global iteration $t$ is bounded by a constant $\tau$. i.e. $\tau_{m, i}^t \le \tau $
        \end{assumption} 

    \subsection{Theorems}
        \begin{theorem} \label{theo:1}
            Under assumption \ref{assum:feasible_optimal_solution} - \ref{assum:bound_delay}, to solve the problem \ref{eq:problem} with algorithm \ref{algo:1} the following inequality holds.
            \begin{align} %\label{eq:result-2}
                            & \frac{1}{T } \sumt \BE \normsq{\nabla \f{w_0^{t}, \bw^{t}} }  \NN
                \leos{}     & \frac{4\ps \BE \bigp{f^0 - f^*}}{T \eta} + \eta \bigp{ 4 \ps \Ls \sigma_*^2 + 8 \ps \Ls \ds \sigma_*^2 + \ps \Ls^3 \mus^2 \ds^2}\NN
                            & + \eta^2\bigp{ 18 \ps \tau^2  \Ls^2 \ds \mGs^2 + 5 \ps \tau^2 \Ls^2 \mus^2 \Ls^2 \ds^2 }  + \mus^2\bigp{\ps \Ls^3 \ds^2}\nonumber
            \end{align}
            where $\Ls = \max_m \bigcb{L, L_0, L_m}$, $ \ds = \max_m \bigcb{ d_{m} } $, $\eta_0 = \eta_m = \eta \le \frac{1}{4 \Ls \ds}$, $\frac{1}{\ps} = \min_m p_m $, $\mus = \max_m \bigcb{\mu_m} $, $\mGs = \max_m \bigcb{\mG_m}$, and $T$ is the number of iterations. 
        \end{theorem}
        
        \begin{remark}
            Theorem~\ref{theo:1} tells that the major factors that affect the convergence are the learning rate $\eta$, the smoothing coefficient $\mu$ for the ZOO, and the biggest parameter size $\ds$ among the clients.  
        \end{remark}
    
        \begin{corollary} \label{coro:1}
            If we choose $\eta = \frac{1}{\sqrt{T}}$, $ \mu = \frac{1}{\sqrt{T}}$, we can derive 
            \begin{align}
                        &   \frac{1}{T} \sumt \BE \normsq{\nabla \f{w_0^{t}, \bw^{t}} }  \NN
            \leos{}     &   \frac{1}{\sqrt{T}} \bigsb{ 4\ps \BE \bigp{f^0 - f^*} +  4 \ps \Ls \sigma_*^2 + 8 \ps \Ls \ds \sigma_*^2 } \NN
                        & + \frac{1}{T}\bigp{ 18 \ps \tau^2  \Ls^2 \ds \mGs^2 + 5 \ps \tau^2 \mus^2 \Ls^4 \ds^2 + \ps  \Ls^3 \ds^2} \NN  
                        & + \frac{1}{T^\frac{3}{2}} \bigp{ \ps \Ls^3 \ds^2 } \nonumber
            \end{align}
            where the parameters are the same as that in Theorem~\ref{theo:1}.
            
        \end{corollary}
    
        \begin{remark}
            Corollary~\ref{coro:1} demonstrates the convergence of our cascaded hybrid optimization framework and shows that it converges in $\mathcal{O}\bigp{\frac{d_*}{\sqrt{T}}} $, where $d_* = \underset{m}{\max} \bigcb{d_{m}}$ represents the largest model size among the clients, and $T$ denotes the number of iterations.
        
            %\begin{align}
            %    \frac{1}{T} \sumt \BE \normsq{\nabla \f{w_0^{t}, \bw^{t}} } = \mathcal{O}\bigp{\frac{d}{\sqrt{T}}} \nonumber
            %\end{align}
        \end{remark}
        
        \begin{remark}
            Comparing our convergence analysis result and ZOO-VFL~\cite{zhang2021desirable}, our result does not include the parameter size of the server ($d_0$) in the constant terms, which demonstrates that the convergence of the global model is not limited by the size of the server's parameter. Therefore, in our framework, the server can apply a larger model without impacting the convergence of the global model. 
        \end{remark}

\section{Security Analysis} \label{sec:security_analysis}
    \subsection{Threat Model}
        We discuss the privacy protection of our framework under the ``honest-but-curious'' and ``honest-but-colluded'' model. 
        \paragraph{Honest-but-curious}
            The "honest but curious" threat model refers to a scenario in which a participant is honest and adheres to the protocol, but is curious about the data of other parties. This party may attempt to gain more knowledge about the data of other parties through communication between participants. Specifically, in VFL, clients seek to infer the label from the server, while the server aims to derive the feature from the client.
                
        \paragraph{Honest-but-colluded}
            The "honest but colluded" threat model involves multiple participants colluding to gain more knowledge about the private data from other participants. Specifically, in VFL, clients may work together to infer the label from the server, or the server may collude with some clients to infer the feature from the remaining clients. 

    \subsection{Theorem}
        \begin{theorem}
            Our framework can defend against existing privacy inference attacks on VFL under the "honest-but-curious" and "honest-but-colluded" scenarios. 
        \end{theorem}

        \noindent\textit{Proof:} \textbf{Defend Against Label Inference Attack:}
            % Server act as black-box to the client. 
            Our framework protects the label on the server by concealing its internal information from clients. 
            Specifically, the server responds to the client with the losses of the model, which are limited to a single value for each batch, without revealing the domain of the target task. Moreover, the server keeps the internal details of its model and the domain information associated with the labels confidential from clients. 
            This approach ensures that the server acts as a black box to clients, who can cooperate with the server without knowing any task information.
            
            In the context of the "honest-but-curious" model, one client in the VFL system attempt to infer the label from the server.
            The ``direct label inference" attack from Fu et al.~\cite{fu2022label} is based on the gradient information provided by the server and relies on strong assumptions about both the attacker and the victim. Specifically, the attack assumes that the server simply sums the output from all clients and that the attacker has explicit knowledge of this fact. By exploiting this information, the label can be directly inferred from the sign of the element in the gradient provided by the server. However, this attack is not feasible for our framework, as we do not transmit gradients to the client and the server model is agnostic, rather than a simple summation. 
            The ``model completion attack'' from Fu et al.~\cite{fu2022label} and the ``forward embeddings leakage'' from Sun et al.\cite{sun2022label} utilize the client's local model and feature to predict the label on the server. In order for these attacks to be successful, the local model and local feature must be well-represented on the target task. Besides, a certain label for the sample cannot be guaranteed with those attacks. Additionally, these attacks assume that the client has knowledge on the target task, which can be avoided by using our proposed framework.
            Deep leakage from gradient and its variant~\cite{zhu2019deep, zhao2020idlg, jin2021cafe} utilize the gradient provided by the server as the optimization objective to reconstruct the true label from a dummy label. However, these attacks assume the attacker has access to the server's model, which is not applicable to our current framework.

            % Honest but colluded. 
            Under the ``honest-but-colluded'' model, some clients collude to infer the label from the server, the attacker can access more information in this scenario. 

            If all clients colluded, the ``direct label inference attack", from Fu et al. \cite{fu2022label} still assume that the client knows that the server uses a simple summation model, which is not applicable to our framework. The ``model completion attack'' from Fu et al. and the ``forward embeddings" attack from Sun et al.\cite{sun2022label} can have better representation on the global task if some client colluded. However, the clients still cannot access the task information from the server, which is not applicable to our model. 
            In the ``honest-but-colluded" model, the deep leakage from the gradient, \cite{zhu2019deep}, still required the gradient information from the server and assumes a simple summation model on the server, which can be avoided with our framework.

        \textbf{Defend Against Feature Inference Attack:}
            Our framework protects the client's features by concealing their internal information from other participants.
            Clients send the model's output for each batch to the server without revealing the feature's domain. 
            Additionally, the server is unable to access the client's model information. 
            As a result, adversaries view the client as a black box, only able to receive outputs from it. This makes it difficult to infer the feature from the client.

            In the "honest-but-curious" model, the server attempts to infer the feature from the clients. 
            % Deep leakage from gradient. 
            The ``deep leakage from gradient''~\cite{zhu2019deep} leverages the gradient as the optimization target to infer the feature from the client. However, this method assumes that the server, as the attacker, can access the client's model, which is not possible through the protocol in our framework. 
            % Model inversion attack 
            The model inversions attack~\cite{fredrikson2015model} uses the model's output to recover the input of a machine-learning model, which has the potential to be used for feature inference attacks in VFL.
            However, this attack requires the attacker to have the ability to adaptively query the target model, which the server does not possess this capability in our framework.

            % Honest but colluded. 
            The "honest-but-colluded" model allows the server to collude with certain clients to infer features from the remaining clients.
            Luo et al.~\cite{luo2021feature} consider a feature inference scenario with two participants, where one participant takes the role of server and client and attempts to infer the feature from the remaining client. 
            They assume that the client uses a logistic regression model, which allows them to reverse the model with the output. 
            However, this method is not applicable to our framework because the client model is agnostic to the attacker. 
            Weng et al.~\cite{weng2020privacy} consider a similar VFL with extra HE scheme, and they assume that the coordinator with the private key also collude, enabling the attacker to decrypt the communication. However, this approach is not applicable to our framework as they also assume a specific model on the client.

\section{Experiments} \label{sec:experiment}
    % The experiment demenstrate what. General introduction. 
    In this section, we did extensive experiments to demonstrate the security of our framework, the convergence of our framework and the feasibility of applying our framework to deep learning tasks.

    \subsection{Experiment Setups}

        % What dataset we use. How we partition the dataset. 
        \paragraph{Datasets}
        % What dataset we use. 
        We vertically partitioned the dataset among $M$ clients, with each client holding an equal amount of features. 
        The server held the labels. Both clients and the server knew the sample IDs, enabling them to coordinate training on each sample. 
        For the base experiment, we used the MNIST dataset~\cite{lecun2010mnist}, the features of the image were flattened and equally distributed among the clients. 
        For image classification with a larger model, we used the CIFAR-10 dataset~\cite{Krizhevsky09learningmultiple}, with each client holding half of each image. 
        For the natural language processing (NLP) task, we used the IMDb dataset~\cite{mcauley2013hidden} where the client held the review text data. 

        \paragraph{Models}
        We used a multilayer perceptron (MLP) for the base experiment to demonstrate the convergence rate of our framework. Although simple, it showed the advantage of our framework. 
        % base model for clients
        The base model for clients was a single-layer Fully Connected Layer (FCL) with an input size equal to the feature size of the client's data and an output size of 128 by default. The activation function was ReLU.
        % base model for the server
        The base model for the server was a two-layer FCL whose input was the concatenation of all the clients' outputs [$c_{i, 1}, \hdots, c_{i, M}$]. Since the client updated asynchronously, the server held a table of [$c_{i, 1}, \hdots, c_{i, M}$]. When the server received an update from client $m$, it would update the corresponding $c_{i, m}$ in the table and use the table as input of the model. The embedding size of the first layer was 128 by default and the output size of the second layer was the number of classes. 
        
        % ResNet 18, model
        For the image classification task, we applied a split ResNet-18 model~\cite{he2016deep} on the VFL framework. 
        There were two clients and one server. 
        Each client held half of each image while the server held the labels. 
        The clients preprocessed the images and passed them through the first convolutional layer of ResNet-18. 
        The model on the server comprised the remaining parts of the ResNet-18 model.
        % NLP 
        For the NLP task, we applied a split distilBERT~\cite{devlin2018bert} model on the VFL framework. 
        The network consisted of one client and one server, the client holding the embedding layer of the transformer and the server holding the remaining parts of the model.

        \paragraph{The Framework for Comparison}
        % the framework we compare. 
        We conducted a comparative analysis of our asynchronous VFL framework with four baseline methods: VAFL~\cite{chen2020vafl}, ZOO-VFL~\cite{zhang2021desirable}, Split-Learning~\cite{vepakomma2018split}, and Syn-ZOO-VFL\footnote{This is the synchronous version of ZOO-VFL and the algorithm is in the Appendix B.}. All baselines employ a single optimization method across the entire VFL, and we applied the same base models to all frameworks.
        While ZOO-VFL and Syn-ZOO-VFL share the same message transmission method as our framework, VAFL and Split-Learning transmit partial derivatives through the network, which poses a privacy risk. It is worth noting that our framework offers the same level of privacy security as ZOO-VFL and Syn-ZOO-VFL, whereas VAFL is privacy risky. Therefore, we consider the experiment on VAFL and Split-Learning as an upper bound for convergence rate comparison among these frameworks, but it is not practical due to the privacy risk.

        \paragraph{Training Procedures}
        % Base model 
        We employed different learning rates for the server and clients in our experiments, as their update times differ. 
        The optimal learning rate $\eta$ was selected from the range [0.020, 0.015, 0.010, 0.005, 0.001] for all frameworks. 
        We chose this range because $\eta=0.001$ was too small, resulting in slow convergence, while $\eta=0.020$ was too large for ZOO to achieve satisfactory test accuracy. 
        We set $\mu$ to 0.001 for all experiments, which was the optimal parameter selected from the range [0.1, 0.01, 0.001, 0.0001, 0.00001] through preliminary experiments.
        To make a fair comparison, we applied the vanilla SGD strategy to all VFL frameworks. The number of training epochs was 100 by default to ensure model convergence. 

        % Training: CIFAR-10
        For training the split ResNet-18 on distributed CIFAR-10, we trained the model for 40 epochs. 
        To determine the optimal learning rate $\eta$ for the framework, we searching $\eta$ within the range [0.03, 0.01 0.003, 0.001] for the framework. 
        We selected the one with the highest test accuracy. 
        For the ZOO-VFL and Syn-ZOO-VFL, we searched for the optimal learning rate in an exponential manner, i.e., [$ \cdots, 3\times 10, 10, 3, 1, 0.3, 0.1, \cdots $]. 
        The upper limit for the searching was where the loss kept increasing, and the lower limit was where the model training accuracy did not increase for every epoch. We selected the learning rate that allowed the model to train the fastest.
        % Training:  NLP
        For the NLP task, we finetuned the pre-trained distil-BERT model. Since the model is pre-trained, we set the number of training epochs to 10. The hyperparameter tuning scheme was the same as that used for the CIFAR-10 task.

    \subsection{A Demonstration on Label Inference Attack}
    
        % Experiment procedure. 
        In this experiment, we aimed to demonstrate the security levels of ZOO-based VFL (ZOO-VFL, Syn-ZOO-VFL and ours) and FOO-based VFL (Split-Learning and VAFL) against a direct label inference attack from Fu et al.~\cite{fu2022label}. 
        The attack is only effective for the ``model without split'' VFLs where the server simply sums up the output from all clients. 
        The threat model involves a curious client aiming to infer labels from the victim server. The client can design the query for the server to acquire partial derivative w.r.t. the global model's output layer, i.e., $\frac{\partial \mathcal{L}(y; y_i) }{\partial y^c}$, where $y$ represents the probability output for all classes, $y^c$ is the probability for the $c$-th class predicted by the model, and there are $C$ classes in total. 
        The label can be directly inferred with the sign of $\frac{\partial \mathcal{L}(y; y_i) }{\partial y^c}$, i.e., if the sign of it is negative, then the label for sample $i$ is $c$; otherwise, the sign is positive. 
        Note that this attack scenario where the server model simply sums the output of the clients is very strong (the server is too vulnerable). However, it has effectively demonstrated the vulnerability of transmitting gradients in VFL.

        % Adversary implement. 
        To simulate a curious client who wanted to infer the label from the server, we designed a dummy client that directly generated a random vector $c_{i, m} \in \mathcal{R}^C$, with elements sampled from $ \mathcal{N}(0, 1)$. 
        The client then randomly selected a $u\in \mathcal{R}^C$ to compute $ \hat{c}_{i, m} = c_{i, m} + u $. 
        The server then responded with the corresponding losses $\hat{h}_{i, m}$ and $h_{i, m}$, and the curious client estimated $\frac{\partial \mathcal{L}(y; y_i) }{\partial y^c}$ using gradient estimation, i.e. $ \hn_{y}\mathcal{L}(y; y_i) = \frac{\phi(d)}{\mu}(\hat{h}_{i, m} - h_{i, m}) u $. 
        In addition to the curious client, eavesdroppers also sought to infer labels from the server. 
        However, when clients are benign, eavesdroppers cannot obtain the client's $u$ value.
        Therefore, in the experiment, they randomly generated a $u$ to estimate the gradient.

        % Data implement. 
        We conducted the label inference attack using the MNIST dataset, using a batch size of 64. 
        The attack success rate was calculated by dividing the number of correctly predicted samples by the total number of samples.
        he VFL framework was run for a single epoch, during which the attacker predicted the label of all samples based on the information they obtained.
        The VFL framework consisted of two clients and one server, where the server model summed up the output from the clients and replied with the losses value w.r.t. the client's output. 
        In the trial involving the curious client, there was one curious client and one benign client. In the trial involving the eavesdropper, both clients were benign. 

        \begin{table}[tbp]
          \centering
          \caption{Demonstration with Direct Label Inference Attack}
          \label{tab:label_inference_attack}
          \begin{tabular}{ccc}
            \toprule
                & FOO frameworks & ZOO frameworks\\
            \midrule
            Curious Client & $100_{\pm 0.0}$ & $11.7_{\pm 0.07} $ \\
            Eavesdropper   & $100_{\pm 0.0}$ & $10.0_{\pm 0.1}$ \\
          \bottomrule
        \end{tabular}
        \end{table}
        
        The results are present in Table~\ref{tab:label_inference_attack}, where each experiment consists of 5 independent trials. The table indicates that the use of FOO in VFL poses a serious privacy vulnerability, as both curious clients and eavesdroppers can infer certain labels. 
        On the other hand, when ZOO is applied to VFL, the malicious client who dedicated designed the query only gained a slight advantage with one query. 
        Additionally, eavesdroppers were unable to infer the label from the messages due to the lack of gradient information on the server. 

    \subsection{The Convergence for Different Numbers of Clients}

        \begin{table*}[!tbp]
            \caption{The Test Accuracy for the Experiments (5 Independent Runs)}
            \label{tab:test_acc}
            \centering
            \begin{tabular}{ccccccccc}
             \toprule
             & \multicolumn{3}{c}{MNIST} & \multicolumn{3}{c}{MNIST} & CIFAR-10 & IMDb \\
             \cmidrule(r){2-4} \cmidrule(r){5-7} \cmidrule(r){8-8} \cmidrule(r){9-9}
             & \multicolumn{3}{c}{MLP - number of clients} & \multicolumn{3}{c}{MLP - server embedding size} & \multirow{2}{*}{ResNet-18} & \multirow{2}{*}{Distil-BERT} \\
             \cmidrule(r){2-2} \cmidrule(r){3-3} \cmidrule(r){4-4}  \cmidrule(r){5-5} \cmidrule(r){6-6} \cmidrule(r){7-7}
             & 4 & 6 & 8 & 128 & 256 & 512 &  &  \\
             \midrule
            Split learning~\cite{vepakomma2018split}   & $97.7_{\pm 0.1} $          & $97.7_{\pm 0.1} $ & $97.5_{\pm 0.2} $ & $97.7_{\pm 0.1} $              & $98.1_{\pm 0.2} $ & $98.1_{\pm 0.1} $ & $84.7_{\pm 0.2}$  & $90.5_{\pm0.1}$ \\
            VAFL~\cite{chen2020vafl}                   & $ 97.7_{\pm 0.2}$          & $ 97.8_{\pm 0.1}$ & $ 97.7_{\pm 0.2}$ & $ 97.7_{\pm 0.2}$ & $ 97.8_{\pm 0.2}$ & $ 97.8_{\pm 0.1}$ & $ 88.1_{\pm 0.1}$ & $ 90.5_{\pm 0.1}$ \\
            \midrule
            Syn-ZOO-VFL                                & $87.4_{\pm 0.3}  $         & $87.4_{\pm 0.2} $ & $87.7_{\pm 0.3} $ & $87.5_{\pm 0.4}  $              & $88.7_{\pm 0.2} $ & $88.2_{\pm 0.3}$ & -                 & \\
            ZOO-VFL\cite{zhang2021desirable}           & $89.0_{\pm 0.3}$           & $89.4_{\pm 0.4}$  & $89.2_{\pm 0.4}$ & $89.0_{\pm 0.3}$ & $85.3_{\pm 0.8}$ & $86.0_{\pm 0.7}$ & - & - \\ 
            VFL-Cascaded & $\textbf{96.4}_{\pm 0.3}$   & $\textbf{96.5}_{\pm 0.4}$ & $\textbf{96.4}_{\pm 0.3}$ & $\textbf{96.4}_{\pm 0.3}$ & $\textbf{96.5}_{\pm 0.4}$ & $\textbf{96.2}_{\pm 0.3}$ & $\textbf{87.2}_{\pm 0.6}$ & $\textbf{89.6}_{\pm 0.2}$ \\
            \bottomrule
            \end{tabular}
            \vspace{-10 pt}
        \end{table*}

        % Add the VAFL
        \begin{figure}[!tbp]
            \centerline{\includegraphics[width=0.9\linewidth]{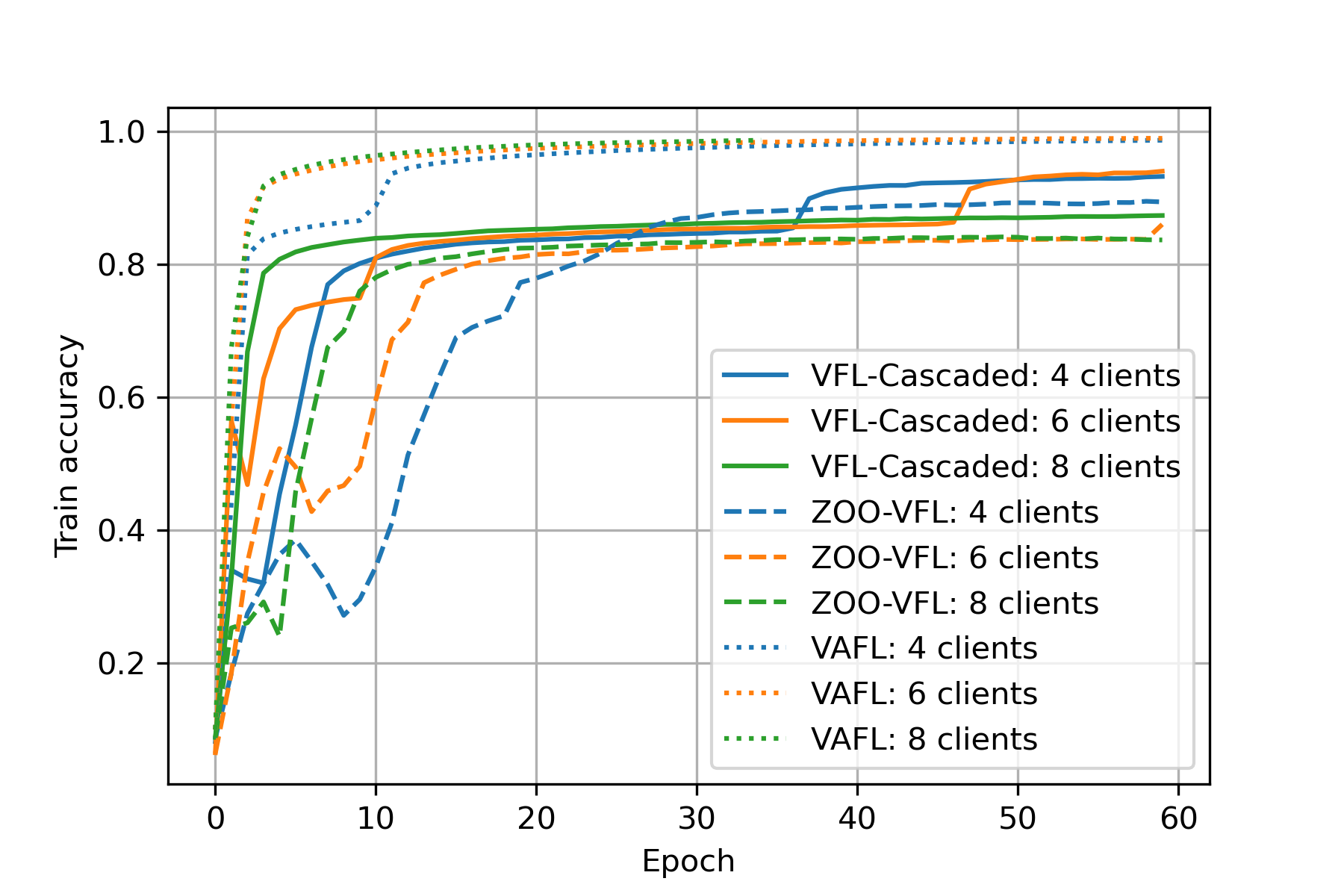}}
            \caption{Learning Curve for Different Numbers of Clients}
            \label{fig:exp_base_n_client}
        \end{figure}
        
        % Set up
        In this experiment, we compared the convergence curve between our framework and others, with varying numbers of clients. With the base model, we set the number of clients to \{4, 6, 8\} and plotted the epoch-training accuracy curve in Figure~\ref{fig:exp_base_n_client}. As illustrated in the figure, our framework exhibited a more stable convergence rate than ZOO-VFL. The curve for ZOO-VFL displayed significant vibration between the fifth and tenth epoch, primarily due to client delay. This phenomenon was less obvious in our framework. 
        Table~\ref{tab:test_acc} shows the test accuracy. Our framework demonstrated a slight test accuracy loss compared to VAFL, which was a trade-off for improving the privacy and security of the framework. In contrast, our framework achieved a much higher test accuracy than ZOO-VFL, indicating that ZOO-VFL does not possess good convergence characteristics.

        \paragraph{More Robust Hyperparameter Tuning}
            \begin{figure}[!tbp]
                \centering
                \centerline{\includegraphics[width=0.9\linewidth]{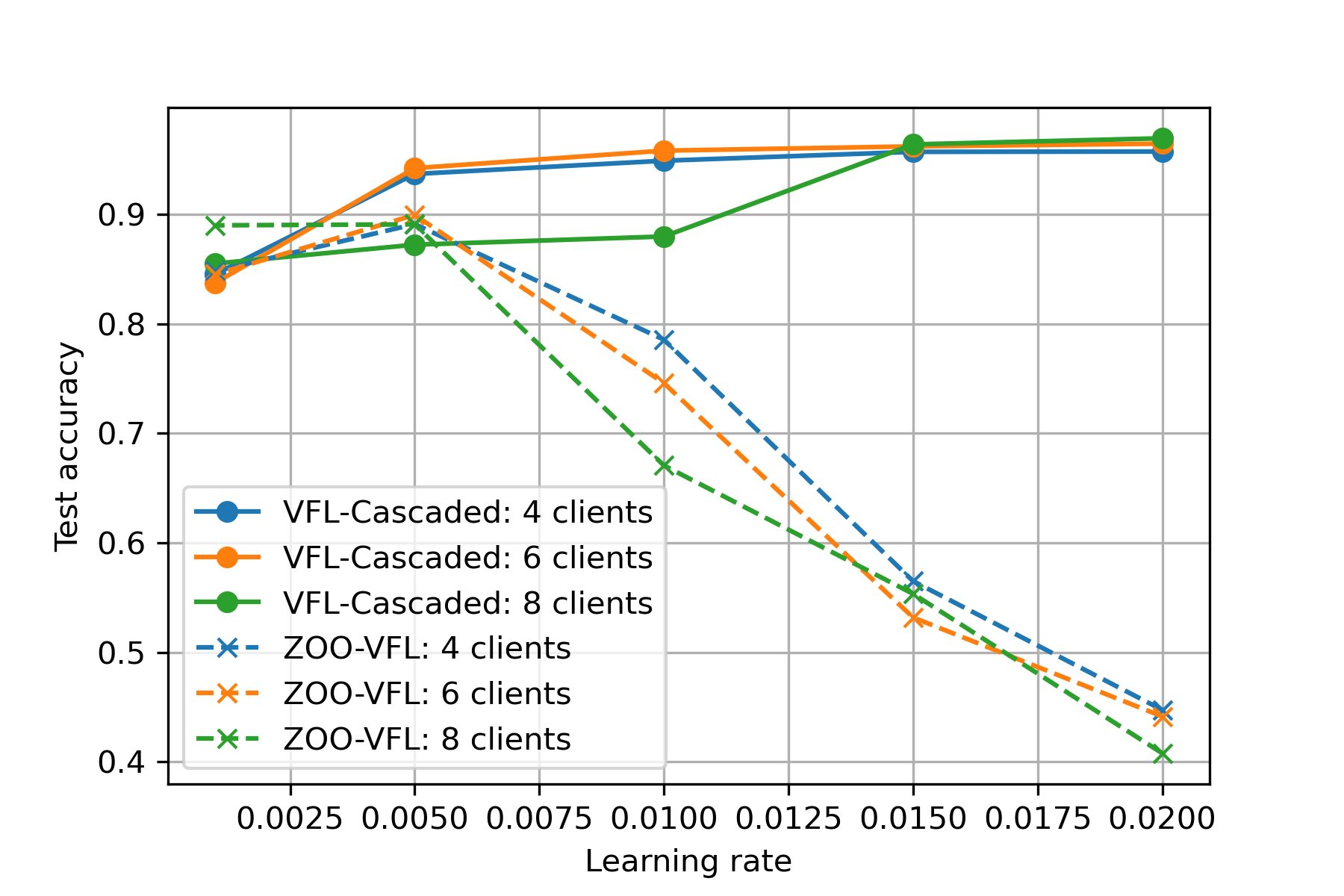}}
                \caption{Robustness of the Hyperparameter}
                \label{fig:hyperparameter_robustness}
            \end{figure}
        
            When searching for the optimal learning rate, we observed that the selection of the learning rate for ZOO-VFL was more sensitive compared to VFL-Cascaded. This sensitivity is an undesirable characteristic for hyperparameter tuning, especially in federated learning, which introduces more hyperparameters than centralized training~\cite{kairouz2019advances}.
            
            % detailed explanation. 
            Assuming that we have obtained the optimal learning rate for ZOO-VFL, it is worth noting that even a slight increase in the learning rate can lead to a significant reduction in test accuracy. Conversely, a minor decrease in the learning rate can also slow the convergence and decrease test accuracy. In contrast, our framework demonstrates greater resilience in learning rate selection, resulting in a more stable performance with less deviation in hyperparameters.
            
            % Experiment result
            To demonstrate the resilience of our framework, we reported the test accuracy at a different learning rate for comparing the ZOO-VFL and VFL-Cascaded. We selected the server learning rate from [0.020, 0.015, 0.010, 0.005, 0.001], and trained the model for 200 epochs to make sure the model converges. The test accuracy is presented in Figure~\ref{fig:hyperparameter_robustness}. Our findings indicate that the deviation from the optimal learning rate had a more significant impact on ZOO-VFL than VFL-Cascaded.
            % Explain the importance. 
            In VFL, a more robust hyperparameter is favorable as it requires less tuning and computational resources. This is particularly important as communication between the server and clients in VFL is costly.
            
    \subsection{The Convergence for Different Server Model Sizes}
        
      % Add the VAFL and change to subfigure. 
      \begin{figure*}[htbp]
          \centering
            \begin{subfigure}[b]{0.32\textwidth}
                 \centering
                 \includegraphics[width=\textwidth]{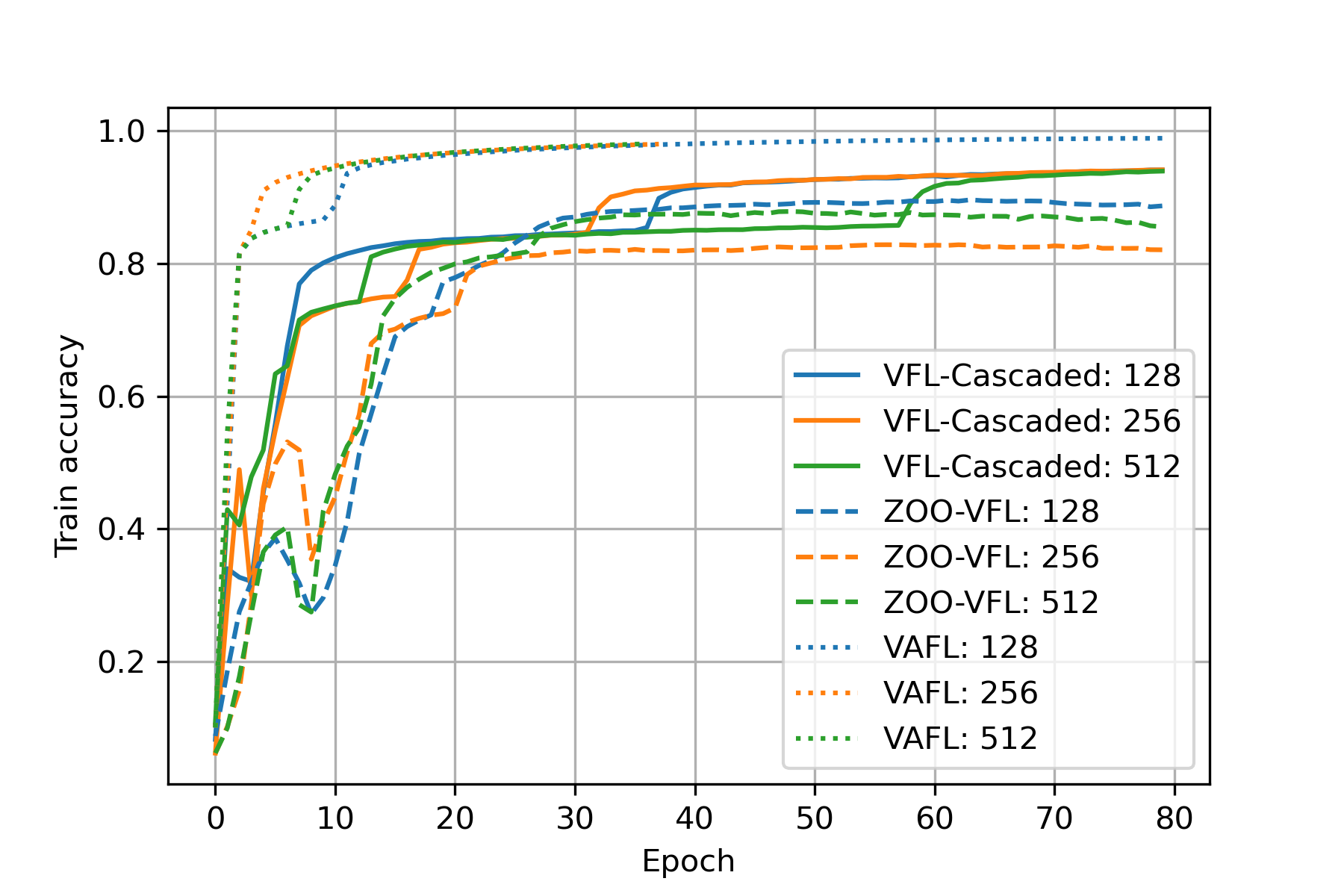}
                 \caption{MNIST - Base Model}
                 %\label{fi
             \end{subfigure}
            \hfill
            \begin{subfigure}[b]{0.32\textwidth}
                \centering
                \includegraphics[width=\textwidth]{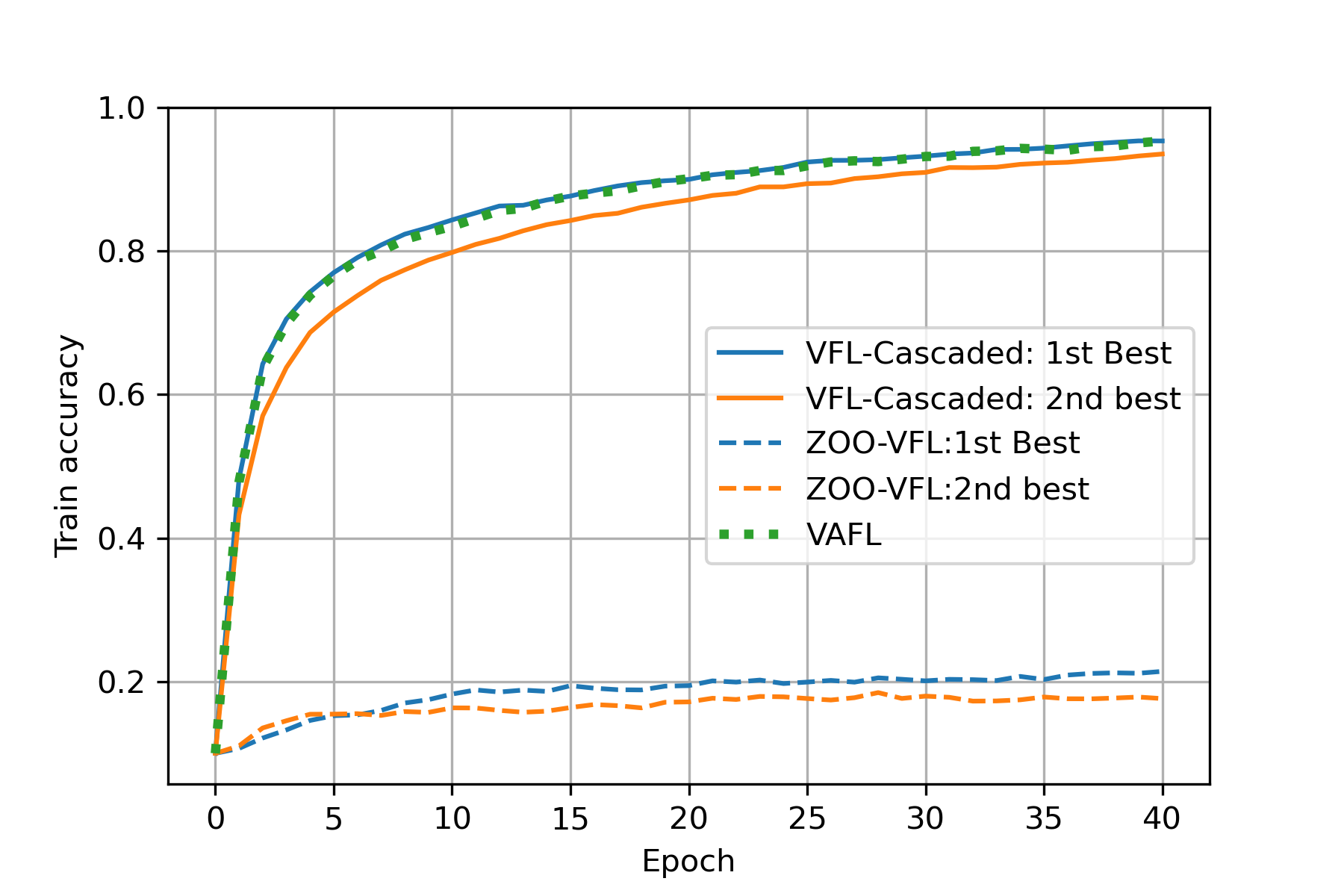}
                \caption{ CIFAR10 - ResNet18 }
            \end{subfigure}
            \hfill
            \begin{subfigure}[b]{0.32\textwidth}
                \centering
                \includegraphics[width=\textwidth]{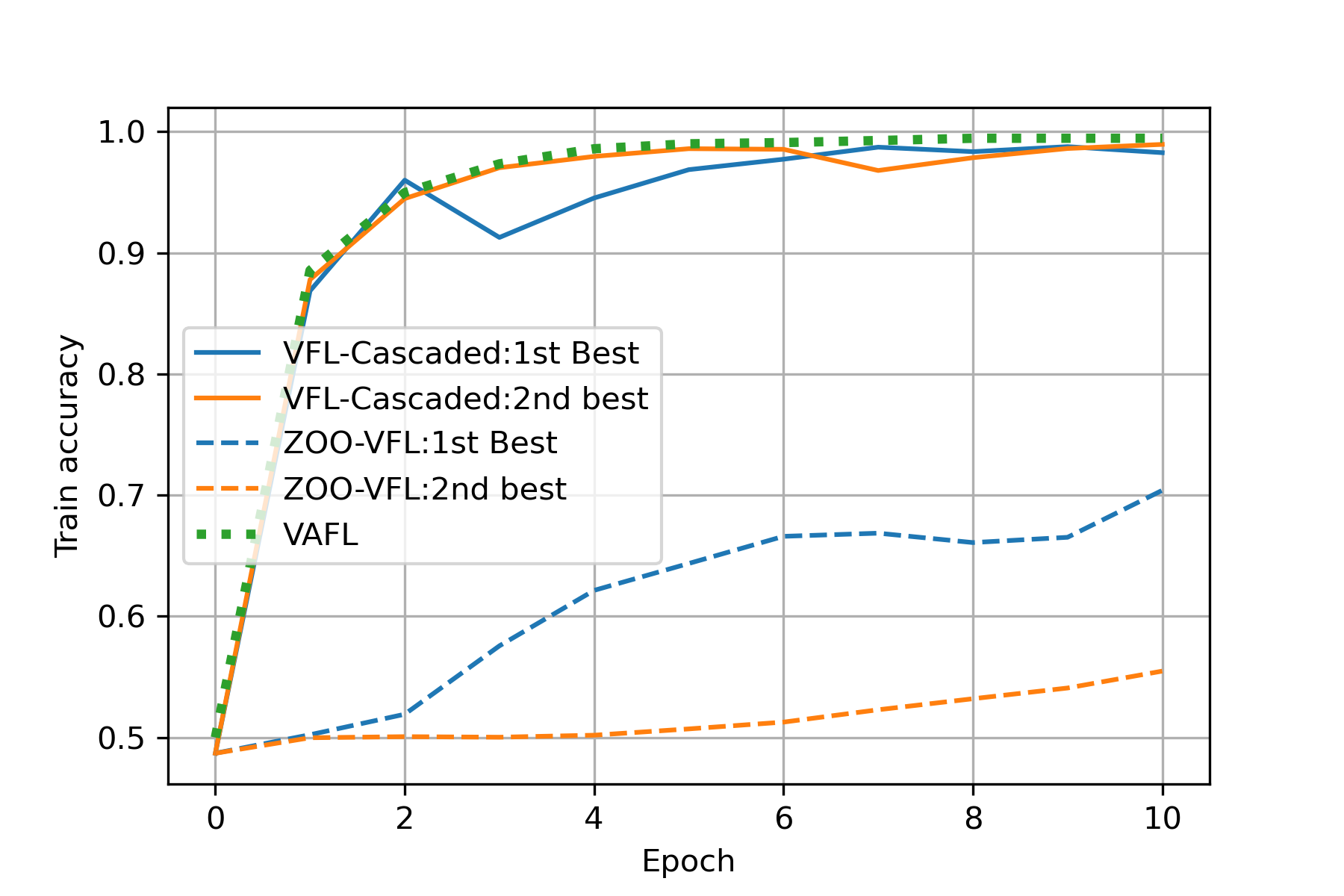}
                \caption{ IMDb - DistillBERT }
            \end{subfigure}
          \caption{Learning Curve for Different Server Model Size}
          \label{fig:exp_server_model_size_all}
      \end{figure*}

        \paragraph{Base Model}
            In this experiment, we conducted a comparison of the convergence rates between our framework and other frameworks, using a variety of server model sizes. 
            The frameworks were applied to four clients and one server, and we tested it on different widths of the server model, specifically the embedding size of the first layer. 
            We varied the embedding size of the first layer of the server from the default value of 128 to 256 and 512, resulting in server model parameter counts of 66954, 133898, and 267786, respectively. 
            % result. 
            The training curve is presented in Figure~\ref{fig:exp_server_model_size_all} (a). As shown in the figure, for all different sizes of models, our framework has a more stable convergence than ZOO-VFL, where the vibration between the fifth and tenth epoch is less obvious. Table~\ref{tab:test_acc} present the test accuracy. For all model sizes, our model has a significant higher test accuracy than ZOO-VFL. However, when compared to VAFL, our framework incurs a trade-off of approximately $1\%$ in test accuracy for privacy security. 

       %\begin{figure}[htbp]
       %    \centering
       %    \centerline{\includegraphics[width=0.9\linewidth]{Figure/exp_model_size.png}}
       %    \caption{The Learning Curve for Different Server Model Sizes}
       %    \label{fig:exp_model_size}
       %\end{figure}

        To demonstrate the superiority of our framework in training larger models, we conducted tests on deep learning tasks, including image classification and text classification (NLP).
        \paragraph{Image Classification}
        % Analysis of the result
        The training curve for the image classification task on CIFAR-10 using the split ResNet-18 model is presented in Figure~\ref{fig:exp_server_model_size_all} (b). As depicted in the figure, our framework maintains a reasonable convergence rate and is robust for the best two learning rates, where the best curve almost overlaps the training curve for VAFL. The training accuracy for ZOO-VFL gradually increases from 0.10 to 0.22 during the training process, indicating the slow convergence problem of ZOO-VFL with the large model. Table~\ref{tab:test_acc} shows the test accuracy. By applying our framework, we can achieve a reasonable test accuracy in 40 training epochs using a modified split ResNet18 model.
        
        %\begin{figure}[htbp]
        %    \centering
        %    \centerline{\includegraphics[width=0.9\linewidth]{Figure/exp_CIFAR.png}}
        %    \caption{The Learning Curve for CIFAR-10 with Split ResNet18}
        %    \label{fig:exp_CIFAR}
        %\end{figure}
        %
        % Finally we get a 88% test accuracy on CIFAR-10. 
        \paragraph{Natural Language Processing}
        We also demonstrated that a more complex transformer-based model for NLP can be trained with our VFL framework. The training curve is depicted in Figure~\ref{fig:exp_server_model_size_all} (c). The dataset comprises of two classes, therefore, the training accuracy commences at around 50\%.

        % Analysis
        The difference in convergence speed becomes more noticeable when using a large model. 
        In our framework, the training accuracy reached 94\% in the second epoch, which took approximately 45 minutes. 
        In contrast, ZOO-VFL's training accuracy only rose from 50\% to 70\% in 10 epochs, requiring around 6 hours of training time, and the model's performance remained close to random guessing.
        Besides, the learning rate was more robust for VFL-Cascaded, with most of the parameters we tuned proving to be effective. 
        In contrast, ZOO-VFL's second-best learning rate exhibited much slower convergence, and the third-best learning rate failed to converge altogether.
        The test accuracy of our model is presented in Table~\ref{tab:test_acc}. 
        Since training for around 6 hours is contrary to the basic idea of fine-tuning, we test the model after 2 epochs of training. 
        The results demonstrate that our framework is capable of training an extremely large deep-learning model.

        %\begin{figure}[htbp]
        %    \centering
        %    \centerline{\includegraphics[width=0.9\linewidth]{Figure/exp_IMDB.png}}
        %    \caption{The learning curve for IMDb dataset with split distil-BERT model}
        %    \label{fig:exp_IMDb}
        %\end{figure}

       %\begin{figure}[htbp]
       %    \centering
       %    \centerline{\includegraphics[width=1\linewidth]{Figure/exp_CIFAR_IMDB.png}}
       %    \caption{The learning curve for CIFAR10 and IMDB}
       %    \label{fig:exp_IMDb}
       %\end{figure}

    % 2 Test Accuracy for models. ( or a big table for all test accuracy. run 5 times. )

%\section{Discussions} \label{sec:discussion}

    %\subsection{Advantages}
    %\subsection{More jobs can be arranged on server}
    %\subsection{Applying FO on server stabilized the training}
    
    %\subsection{Limitations}

\section{Limitations and Discussions} \label{sec:discussion}

    \begin{table}[tbp]
      \caption{Comparison with Typical VFL Frameworks } % (Ours is more suitable for the real-world application scenario)
      \label{tab:detail_comparison}
      \begin{tabular}{cccccccccc}
        \toprule
            & \multicolumn{3}{c}{VAFL} & \multicolumn{3}{c}{ZOO-VFL} & \multicolumn{3}{c}{Ours}\\
            \cmidrule(r){2-4} \cmidrule(r){5-7} \cmidrule(r){8-10}
                            & S & C & \textbf{F} & S & C & \textbf{F} & S & C & \textbf{F}  \\
        \midrule
        Model Applicability & \xmark & \xmark & \xmark & \cmark & \cmark & \cmark & \xmark & \cmark & \cmark \\ 
        Fast Convergence    & \cmark & \cmark & \cmark & \xmark & \xmark & \xmark & \cmark & \xmark & \cmark \\ 
        Privacy Security    & \xmark & \xmark & \xmark & \cmark & \cmark & \cmark & \cmark & \cmark & \cmark \\ 
        %\midrule
        Comp. Efficiency    & \cmark & \cmark & \cmark & \cmark & \cmark & \cmark & \cmark & \cmark & \cmark \\ 
      \bottomrule
    \end{tabular}
    \vspace{-10 pt}
    \end{table}

    % still limited by the size of the client model. 
    % To retain the benefit of applying ZOO on VFL, we can only apply the FOO on the server, i.e. we did not boost the entire VFL framework. We boost VFL by carefully utilizing the specialized part of ZOO and FOO, using ZOO to achieve maximum model applicability, privacy-preserving, and communication efficiency, and using FOO on the server for boosting the convergence speed. The convergence rate of the global model is still limited by the size of the client model. Therefore, we can only apply a small client model, e.g. in our experiment, only a few layers of ResNet-18 and distillBERT were applied on the client side. The limitation on the size of the client's model may limit the application of our framework. However, the server, which is the initiator of the VFL, normally takes more responsibility for training the model, therefore our framework is matching with the demand and still has a great prospect for application.

    % still limited by the size of the client model. 
    In our framework, we utilized ZOO and FOO strategically to address the demanding aspects of the VFL framework. Specifically, we employed ZOO on the client to maximize model applicability and privacy protection, and FOO on the server to accelerate convergence for large models. 
    We carefully balanced the advantages and disadvantages of ZOO and FOO in different parts of the VFL model to ensure that our framework meets all requirements for practical VFL.
    A detailed comparison of the frameworks is presented in Table~\ref{tab:detail_comparison} (``S'' for the server and ``C'' for the client, ``\textbf{F}'' for the entire framework). 
    It is important to note that the inherent limitations of ZOO and FOO were not eliminated. That is, ZOO's slow convergence makes it unsuitable for dealing with large models on the client side, while the server can only handle differentiable models.
    
    % Explanation. 
    However, our framework is more suitable for real-world application scenarios for several reasons. 
    Firstly, in VFL, the server is the initiator and sole beneficiary of the framework, with all clients acting as collaborators. As such, it is more cost-effective for the server to train a larger model to achieve better prediction results, as only the server receives the prediction.
    Secondly, the server typically has more computational resources than the clients, making it computationally efficient for the server to train a larger model.
    Thirdly, as the server is the initiator and has the ability to select its model, the model applicability of the server is not as critical in VFL. 
    Conversely, for clients, their models are unknown to the initiator of the VFL, making the model-agnostic characteristic important. Therefore, our framework is more suitable for real-world applications than other frameworks that use a unified optimization method.

    % \begin{table}[t!]
    %   \caption{Comparison with Typical VFL Frameworks (Ours is more suitable for the real-world application scenario)}
    %   \label{tab:detail_comparison}
    %   \begin{tabular}{ccccccc}
    %     \toprule
    %         & \multicolumn{2}{c}{VFL-FOO} & \multicolumn{2}{c}{VFL-ZOO} & \multicolumn{2}{c}{Ours}\\
    %         \cmidrule(r){2-3} \cmidrule(r){4-5} \cmidrule(r){6-7}
    %                         & S & C & S & C & S & C  \\
    %     \midrule
    %     Model Applicability & \xmark & \xmark & \cmark & \cmark & \xmark & \cmark  \\ 
    %     Fast Convergence    & \cmark & \cmark & \xmark & \xmark & \cmark & \xmark  \\ 
    %     Privacy Security    & \xmark & \xmark & \cmark & \cmark & \cmark & \cmark  \\ 
    %     Comp. Efficiency    & \cmark & \cmark & \cmark & \cmark & \cmark & \cmark  \\ 
    %   \bottomrule
    % \end{tabular}
    % \end{table}

\section{Conclusions} \label{sec:conclusion}
    % Conclusion. 
    We proposed a novel VFL framework where different optimization methods were applied to the upstream (server) and the downstream (client) of the VFL cascaded.
    This approach maximized the benefits of both optimization methods. The clients are optimized with ZOO to protect privacy, while the server is optimized with FOO to accelerate convergence without compromising the framework's privacy.
    Theoretical results demonstrated that our framework with cascaded hybrid optimization converges faster than the ZOO-based VFL, and that applying a large model on the server does not hinder convergence.
    Extensive experiments demonstrated that our framework achieves better convergence characteristics compared with the ZOO-based VFL while maintaining the same level of privacy security.

\iffalse
    \section*{Acknowledgments}
\fi


\bibliographystyle{plain}
\bibliography{AC_1_Reference}

\newpage

\onecolumn
\appendices

\section{Convergence Analysis}
    \subsection{notation}
        Below is a notation table for the parameter used in the convergence analysis.
        \begin{table}[H]
            \centering
            \renewcommand{\arraystretch}{1.0}
            \caption{Notation Table}
            \begin{tabular}{ll}
            \toprule
            \multicolumn{2}{c}{}\\
            \multicolumn{2}{c}{\underline{Basic:}}\\
            \multicolumn{2}{c}{}\\
            $ w_0 $ & The parameter for the server. \\
            $ w_m $ & The parameter for the client $m$. \\
            $ \bw = [w_1, w_2, \cdots, w_M] $ & The grouped parameters for all the clients. \\
            $ \f{w_0, \bw} = \f{w_0, \bw, X, y} $ & The global loss function \\
            $ \f[i]{w_0, \bw} = \f[i]{w_0, w_1, \hdots, w_M } $ & The loss function for the sample $i$. \\ 
            \multicolumn{2}{c}{}\\
            \multicolumn{2}{c}{\underline{Notation with timestep ($t$), clients' delay ($\Tilde{\bw}$), ZOO gradient estimator ($\hn$) }}\\
            \multicolumn{2}{c}{}\\
            $ w_{m}^{t} $ & The client $m$'s parameter, at global timestep $t$, \\ 
            % client delay
            $ \bw^t = [w_1^{t}, \hdots, w_M^{t} ] $ & The clients' parameter at global timestep $t$\\
            $ \tbw^t = {{{\bf w}}}^{t-\tau^{t}_{i}} = [w_1^{t- \tau^t_{1, i}}, \hdots, w_M^{t-\tau^t_{M, i}} ] $ & The delayed parameter for all the clients at \\
                                                                                                                  & global time step $t$ (and the local timestep is $0$ for all $w$). \\
            % ZO 
            $ \hn_{w_m} \f[i]{w_0, \bw} = \frac{\phi(d_{h_m})}{\mu_m} \bigsb{  f_i\bigp{w_{m} + \mu_m u_{m, i} }  - f_i\bigp{w_{m}}} u_{m, i} $ & The ZO gradient estimator w.r.t. the client $m$'s parameter $w_m$ \\
            %\multicolumn{2}{c}{}\\
            %\multicolumn{2}{c}{\underline{Cost functions:}}\\
            %\multicolumn{2}{c}{}\\
            \multicolumn{2}{c}{}\\
            \bottomrule
            \end{tabular}
        \label{tab:Notations}
        \end{table}
    %\vfill
    %\newpage

    \subsection{Lemmas }
        %%%%%%%%% Lemma : ZO %%%%%%%%%
        \begin{lemma} \label{lemma:ZO}
            \textbf{Zeroth-Order Optimization.} For arbitrary $f \in C_{L}^{1} (\mathcal{R}^d )$, we have:
            
            1) $f_{\mu}(x)$ is continuously differentiable, its gradient is Lipschitz continuous with $L_\mu \le L$:
            \begin{align} 
                \nabla f_\mu \bigp{x} =  \BE[\bu] \bigsb{ \hn f(x) }  \label{eq:lemma-ZO-1}
            \end{align}
            where $\mathbf u$ is drawn from the uniform distribution over the unit Euclidean sphere, and $\hn f(x) = \frac{d}{\mu} \bigsb{ f \bigp{x + \mu \bu } - f\bigp{x } } u$ is the gradient estimator, $f_\mu(x) = \BE[\bu] \bigsb{ f(x+\mu \bu) }$ is the smooth approximation of $f$.
            
            2) For any $x \in \mathbb R^{d}$,
            \begin{align}
                & | f_\mu \bigp{x} - \f{x} | \le \frac{L {\mu^2}}{2} \label{eq:lemma-ZO-2-1} \\
                & \norm{ \nabla f_{\mu} (x) - \nabla f(x) }^2 \le \frac{{\mu}^2 L^2 d^2}{4} \label{eq:lemma-ZO-2-2} \\
                & \frac{1}{2} \norm{ \nabla \f{x} }^2 - \frac{\mu^2 L^2 d^2}{4} \le \norm{ \nabla f_\mu \bigp{x} }^2 \le 2 \normsq{\nabla\f{x} } + \frac{\mu^2 L^2 d^2}{2} \label{eq:lemma-ZO-2-3}
            \end{align}
    
            3) For any $x \in \mathbb R^{d}$,
            \begin{align} 
                \BE[\bu] \bigsb{ \normsq{\hn \f{x} }  } \le 2 d \normsq{ \nabla \f{x} } + \frac{\mu^2 L^2 d^2 }{2} \label{eq:lemma-ZO-3}
            \end{align}
            % Comment the not used one
            \iffalse
            \begin{align}
                \BE[\bu] \bigsb{\normsq{ \hn \f{x} - \nabla f_\mu \bigp{x} } } \le \BE[\bu] \bigsb{ \normsq{\hn \f{x} }  } \le 2 d \normsq{ \nabla \f{x} } + \frac{\mu^2 L^2 d^2 }{2}
            \end{align}
            \fi
        \end{lemma}
        
        Lemma~\ref{lemma:ZO} helps build a connection between $\f{\cdot}$ and its smooth approximation $ \fmum{\cdot} $ of the convergence analysis. Proof of this lemma is provided in \cite{liu2018zeroth, gao2018information}.

% Bound one global round. (Server and client. )
\subsection{Bound the global update round} \label{appendix:subsec:bound_the_global_update}
    In one global round during training, the client $\mt$ is activated, and the server and the client $\mt$ update one step. 
    %%%%%%%%%%%%%%%%%%%%%%%%%%%%%%%%%%%%%%%%%%%%%%%%%%%%%%%%%%%%%%%%%%%%%%%%%%%% The main, global round. %%%%%%%%%%%%%%%%%%%%%%%%%%%%%%%%%%%%%%%%%%%%%%%%%%%%%%%%%%%%%%%%%%%%%%%%%%%%%%%%%%%%%%%%%%%%%
    Taking expectations w.r.t. the sample $i$ and the random direction $u$ for the zeroth-order optimization in one global update round. 
    \begin{align} \label{eq:global_step}
                  & \BE[i, u] \bigsb{ \f{w_0^{t+1}, w_1^{t}, \cdots,  w_\mt^{t+1}, \cdots, w_M^{t}}  -  \f{w_0^{t}, w_1^{t}, \cdots,  w_\mt^{t}, \cdots, w_M^{t}}} \NN
        \leos{1)} & \uba{ -\eta_0 \BE[i] \ip{\nz \f{w_0^{t}, \bw^{t}} }{ \nz \f[i]{w_0^{t}, \tbw^t } }}\NN
                  & \ubb{ +\frac{1}{2} L \eta_0^2 \BE[i] \normsq{ \nz \f[i]{w_0^{t}, \tbw^t} } }\NN
                  & \ubc{ -\eta_\mt \BE[i, u] \ip{\nmt \f{w_0^{t}, \bw^{t}} }{ \hn_\mt \f[i]{w_0^{t}, \tbw^t}}} \NN 
                  & \ubd{ +\frac{1}{2} L \eta_\mt^2 \BE[i, u] \normsq{\hn_\mt \f[i]{w_0^{t}, \tbw^t} } }\NN
        \leos{2)} & - \frac{1}{2} \eta_0 \BE[i] \normsq{\nz \f{w_0^{t}, \bw^{t}} } +  \frac{1}{2} \eta_0 L_0^2  \BE[i]\normsq{ \bw^t - \tbw^t }  \NN
                  & + L \eta_0^2 L_0^2 \BE[i] \normsq{ \bw - \tbw} + L \eta_0^2 \BE[i]\normsq{\nz \f{w_0^{t}, \bw^t}}  + L \eta_0^2\sigma_0^2 \NN
                  & - \frac{1}{2}\eta_\mt \BE[i, u] \normsq{\nmt \f{w_0^{t}, \bw^{t}} } + \frac{1}{4} \eta_\mt \mu_\mt^2 L_\mt^2 d_\mt^2 + \eta_\mt L_\mt^2 \BE[i, u] \normsq{ \bw^t - \tbw^t} \NN
                  & + 2 L \eta_\mt^2 d_\mt \BE[i, u] \normsq{ \nmt \f{w_0^{t}, \bw^t}} + 2 L \eta_\mt^2 d_\mt L_\mt^2 \BE[i, u] \normsq{ \bw^t - \tbw^t }  + 2 L \eta_\mt^2 d_\mt \sigma_\mt^2 + \frac{1}{4} L \eta_\mt^2 \mu_\mt^2 L_\mt^2 d_\mt^2  \NN
        \leos{2)} & - \bigp{\frac{1}{2} \eta_0 - L \eta_0^2 }\BE[i] \normsq{\nz \f{w_0^{t}, \bw^{t}} } \NN
                  & + \bigp{ \frac{1}{2} \eta_0 + L \eta_0^2}L_0^2 \BE[i] \normsq{\bw - \tbw}  + L \eta_0^2\sigma_0^2 \NN
                  & - \bigp{\frac{1}{2} \eta_\mt - 2 L \eta_\mt^2 d_\mt } \BE[i, u] \normsq{\nmt \f{w_0^{t}, \bw^{t}} } + \bigp{\eta_\mt + 2 L \eta_\mt^2 d_\mt}L_\mt^2  \BE[i, u] \normsq{ \bw^t - \tbw^t} \NN
                  & + 2 L \eta_\mt^2 d_\mt \sigma_\mt^2 + \frac{1}{4} \bigp{ L \eta_\mt^2 + \eta_\mt } \mu_\mt^2 L_\mt^2 d_\mt^2  \NN
        \leos{3)} & - \bigp{\frac{1}{2} \eta_0 - L \eta_0^2 }\BE[i] \normsq{\nz \f{w_0^{t}, \bw^{t}} } - \bigp{\frac{1}{2} \eta_\mt - 2 L \eta_\mt^2 d_\mt } \BE[i, u] \normsq{\nmt \f{w_0^{t}, \bw^{t}} } \NN
                  & + \bigsb{\bigp{ \frac{1}{2} \eta_0 + L \eta_0^2}L_0^2 + \bigp{\eta_\mt + 2 L \eta_\mt^2 d_\mt}L_\mt^2 } \BE[i] \normsq{\bw - \tbw}   \NN
                  & + L \eta_0^2\sigma_0^2 + 2 L \eta_\mt^2 d_\mt \sigma_\mt^2 + \frac{1}{4} \bigp{ L \eta_\mt^2 + \eta_\mt } \mu_\mt^2 L_\mt^2 d_\mt^2 \NN
    \end{align}
    where 1) applies assumption~\ref{assum:smoothness} (smoothness),
    2) plugging in a, b, c\& d, 
    3) collect the equation. 

    %%%%%%%%%%%%%%%%%%%%%%%%%% a) %%%%%%%%%%%%%%%%%%%%%%%%%%
    For a)
    \begin{align}
                  & -\eta_0 \BE[i] \ip{\nz \f{w_0^{t}, \bw^{t}} }{ \nz \f[i]{w_0^{t}, \tbw^t } } \NN
        \eq       & -\eta_0 \BE[i] \ip{\nz \f{w_0^{t}, \bw^{t}} }{ \nz \f[i]{w_0^{t}, \tbw^t } - \nz \f[i]{w_0^{t}, \bw^t } + \nz \f[i]{w_0^{t}, \bw^t } } \NN
        \eq       & -\eta_0 \BE[i] \ip{\nz \f{w_0^{t}, \bw^{t}} }{ \nz \f[i]{w_0^{t}, \tbw^t } - \nz \f[i]{w_0^{t}, \bw^t }} -\eta_0 \BE[i] \ip{\nz \f{w_0^{t}, \bw^{t}} }{\nz \f[i]{w_0^{t}, \bw^t } } \NN
        \eqos{1)} & -\eta_0 \BE[i] \ip{\nz \f{w_0^{t}, \bw^{t}} }{ \nz \f[i]{w_0^{t}, \tbw^t } - \nz \f[i]{w_0^{t}, \bw^t }} - \eta_0 \BE[i] \norm{\nz \f{w_0^{t}, \bw^{t}}}^2  \NN
        \eqos{2)} & - \frac{1}{2} \eta_0 \BE[i] \normsq{\nz \f{w_0^{t}, \bw^{t}} } +  \frac{1}{2} \eta_0 \BE[i]\normsq{ \nz \f[i]{w_0^{t}, \tbw^t } - \nz \f[i]{w_0^{t}, \bw^t }}  \NN
        \eqos{3)} & - \frac{1}{2} \eta_0 \BE[i] \normsq{\nz \f{w_0^{t}, \bw^{t}} } +  \frac{1}{2} \eta_0 L_0^2  \BE[i]\normsq{ \bw^t - \tbw^t }  \NN
    \end{align}
    where 1) applies assumption~\ref{assum:unbiased_gradient} (unbiased gradient), 
    2) applies $\ip{a}{b} \le \frac{1}{2} \norm{a}^2 + \frac{1}{2} \norm{b}^2$, 
    3) applies assumption~\ref{assum:smoothness} (smoothness). 
    % done
    
    %%%%%%%%%%%%%%%%%%%%%%%%%% b) %%%%%%%%%%%%%%%%%%%%%%%%%%
    For b):
    \begin{align}
                    &  \frac{1}{2} L \eta_0^2 \BE[i] \normsq{ \nz \f[i]{w_0^{t}, \tbw^t} } \NN
        \eq         &  \frac{1}{2} L \eta_0^2 \BE[i] \normsq{ \nz \f[i]{w_0^{t}, \tbw^t} - \nz \f[i]{w_0^{t}, \bw^t} + \nz \f[i]{w_0^{t}, \bw^t}  } \NN
        \leos{1)}   &  L \eta_0^2 \BE[i] \normsq{ \nz \f[i]{w_0^{t}, \tbw^t} - \nz \f[i]{w_0^{t}, \bw^t}} + L \eta_0^2 \BE[i] \normsq{\nz \f[i]{w_0^{t}, \bw^t}  } \NN
        \leos{2)}   &  L \eta_0^2 L_0^2 \BE[i] \normsq{ \bw - \tbw} + L \eta_0^2 \BE[i] \normsq{\nz \f[i]{w_0^{t}, \bw^t}  } \NN
        \leos{3)}   &  L \eta_0^2 L_0^2 \BE[i] \normsq{ \bw - \tbw} + L \eta_0^2 \bigp{ \normsq{\nz \f{w_0^{t}, \bw^t}}  + \sigma_0^2 }\NN
        \le         &  L \eta_0^2 L_0^2 \BE[i] \normsq{ \bw - \tbw} + L \eta_0^2 \normsq{\nz \f{w_0^{t}, \bw^t}}  + L \eta_0^2\sigma_0^2 \NN
    \end{align}
    where 1): $\normsq{a+b} \le 2\normsq{a} + 2\normsq{b}$, 
    2) applies assumption~\ref{assum:smoothness} (smoothness), 
    3) applies $\BE(X^2) = \BE(X)^2 + \Var(X)$ and assumption~\ref{assum:bounded_variance} (bounded variance).
    % done
    
    %%%%%%%%%%%%%%%%%%%%%%%%%% c) %%%%%%%%%%%%%%%%%%%%%%%%%%
    For c):
    \begin{align}
                    & -\eta_\mt \BE[i, u] \ip{\nmt \f{w_0^{t}, \bw^{t}} }{ \hn_\mt \f[i]{w_0^{t}, \tbw^t}} \NN
        \eqos{1)}   & -\eta_\mt \BE[i, u] \ip{\nmt \f{w_0^{t}, \bw^{t}} }{ \nmt \f[\mu_\mt, i]{w_0^{t}, \tbw^t}} \NN
        \eq         & -\eta_\mt \BE[i, u] \ip{\nmt \f{w_0^{t}, \bw^{t}} }{ \nmt \f[\mu_\mt, i]{w_0^{t}, \tbw^t} - \nmt \f[i]{w_0^{t}, \bw^t} + \nmt \f[i]{w_0^{t}, \bw^t}} \NN
        \eqos{2)}   & -\eta_\mt \BE[i, u] \ip{\nmt \f{w_0^{t}, \bw^{t}} }{ \nmt \f[\mu_\mt, i]{w_0^{t}, \tbw^t} - \nmt \f[i]{w_0^{t}, \bw^t}} -\eta_\mt \BE[i, u]\normsq{\nmt \f{w_0^{t}, \bw^t}} \NN
        \eqos{3)}   & -\frac{1}{2}\eta_\mt \BE[i, u] \normsq{\nmt \f{w_0^{t}, \bw^{t}} } + \frac{1}{2}\eta_\mt \BE[i, u] \normsq{ \nmt \f[\mu_\mt, i]{w_0^{t}, \tbw^t} - \nmt \f[i]{w_0^{t}, \bw^t}} \NN
        \eqos{}     & -\frac{1}{2}\eta_\mt \BE[i, u] \normsq{\nmt \f{w_0^{t}, \bw^{t}} } \NN
                    & +\frac{1}{2}\eta_\mt \BE[i, u] \normsq{ \nmt \f[\mu_\mt, i]{w_0^{t}, \tbw^t} - \nmt \f[i]{w_0^{t}, \tbw^t} + \nmt \f[i]{w_0^{t}, \tbw^t} - \nmt \f[i]{w_0^{t}, \bw^t}} \NN
        \eqos{4)}   & -\frac{1}{2}\eta_\mt \BE[i, u] \normsq{\nmt \f{w_0^{t}, \bw^{t}} } + \eta_\mt \BE[i, u] \normsq{ \nmt \f[\mu_\mt, i]{w_0^{t}, \tbw^t} - \nmt \f[i]{w_0^{t}, \tbw^t}} \NN % 
                    &  + \eta_\mt \BE[i, u] \normsq{ \nmt \f[i]{w_0^{t}, \tbw^t} - \nmt \f[i]{w_0^{t}, \bw^t}} \NN
        \eqos{5)}   & -\frac{1}{2}\eta_\mt \BE[i, u] \normsq{\nmt \f{w_0^{t}, \bw^{t}} } + \frac{1}{4} \eta_\mt \mu_\mt^2 L_\mt^2 d_\mt^2 + \eta_\mt \BE[i, u] \normsq{ \nmt \f[i]{w_0^{t}, \tbw^t} - \nmt \f[i]{w_0^{t}, \bw^t}} \NN
        \eqos{6)}   & -\frac{1}{2}\eta_\mt \BE[i, u] \normsq{\nmt \f{w_0^{t}, \bw^{t}} } + \frac{1}{4} \eta_\mt \mu_\mt^2 L_\mt^2 d_\mt^2 + \eta_\mt L_\mt^2 \BE[i, u] \normsq{ \bw^t - \tbw^t} \NN
    \end{align}
    where 1) applies Eq.~\ref{eq:lemma-ZO-1} in lemma~\ref{lemma:ZO}, 
    2) applies assumption~\ref{assum:unbiased_gradient} (unbiased gradient),
    3) applies $\ip{a}{b} \le \frac{1}{2} \norm{a}^2 + \frac{1}{2} \norm{b}^2$, 
    4) applies $\normsq{a+b} \le 2\normsq{a} + 2\normsq{b}$, 
    5) applies Eq.~\ref{eq:lemma-ZO-2-2} in lemma~\ref{lemma:ZO}, 
    6) applies assumption~\ref{assum:smoothness} (smoothness).

    %%%%%%%%%%%%%%%%%%%%%%%%%% d) %%%%%%%%%%%%%%%%%%%%%%%%%%
    For d):
    \begin{align} \label{eq:d}
                    & \frac{1}{2} L \eta_\mt^2 \BE[i, u] \normsq{\hn_\mt \f[i]{w_0^{t}, \tbw^t}  } \NN
        \leos{1)}   & \frac{1}{2} L \eta_\mt^2 \BE[i, u] \bigp{ 2d_\mt \normsq{ \nmt \f[i]{w_0^{t}, \tbw^t}} + \frac{1}{2} \mu_\mt^2 L_\mt^2 d_\mt^2 } \NN
        \eqos{  }   & L \eta_\mt^2 d_\mt  \BE[i, u] \normsq{ \nmt \f[i]{w_0^{t}, \tbw^t}} + \frac{1}{4} L \eta_\mt^2 \mu_\mt^2 L_\mt^2 d_\mt^2  \NN
        \eqos{}     & L \eta_\mt^2 d_\mt  \BE[i, u] \normsq{ \nmt \f[i]{w_0^{t}, \tbw^t} - \nmt \f[i]{w_0^{t}, \bw^t} + \nmt \f[i]{w_0^{t}, \bw^t} } + \frac{1}{4} L \eta_\mt^2 \mu_\mt^2 L_\mt^2 d_\mt^2  \NN
        \leos{2)}   & 2 L \eta_\mt^2 d_\mt  \BE[i, u] \normsq{ \nmt \f[i]{w_0^{t}, \tbw^t} - \nmt \f[i]{w_0^{t}, \bw^t}} + 2 L \eta_\mt^2 d_\mt  \BE[i, u] \normsq{ \nmt \f[i]{w_0^{t}, \bw^t} } + \frac{1}{4} L \eta_\mt^2 \mu_\mt^2 L_\mt^2 d_\mt^2  \NN
        \leos{3)}   & 2 L \eta_\mt^2 d_\mt L_\mt^2 \BE[i, u] \normsq{ \bw^t - \tbw^t } + 2 L \eta_\mt^2 d_\mt \BE[i, u] \normsq{ \nmt \f[i]{w_0^{t}, \bw^t} } + \frac{1}{4} L \eta_\mt^2 \mu_\mt^2 L_\mt^2 d_\mt^2  \NN
        \leos{4)}   & 2 L \eta_\mt^2 d_\mt L_\mt^2 \BE[i, u] \normsq{ \bw^t - \tbw^t } + 2 L \eta_\mt^2 d_\mt \bigp{\BE[i, u] \normsq{ \nmt \f{w_0^{t}, \bw^t}} + \sigma_\mt^2 }+ \frac{1}{4} L \eta_\mt^2 \mu_\mt^2 L_\mt^2 d_\mt^2  \NN
        \eqos{  }   & 2 L \eta_\mt^2 d_\mt L_\mt^2 \BE[i, u] \normsq{ \bw^t - \tbw^t } + 2 L \eta_\mt^2 d_\mt \BE[i, u] \normsq{ \nmt \f{w_0^{t}, \bw^t}} + 2 L \eta_\mt^2 d_\mt \sigma_\mt^2 + \frac{1}{4} L \eta_\mt^2 \mu_\mt^2 L_\mt^2 d_\mt^2  \NN
        \eqos{  }   & 2 L \eta_\mt^2 d_\mt \BE[i, u] \normsq{ \nmt \f{w_0^{t}, \bw^t}} + 2 L \eta_\mt^2 d_\mt L_\mt^2 \BE[i, u] \normsq{ \bw^t - \tbw^t }  + 2 L \eta_\mt^2 d_\mt \sigma_\mt^2 + \frac{1}{4} L \eta_\mt^2 \mu_\mt^2 L_\mt^2 d_\mt^2  \NN
    \end{align}     
    where 1) applies Eq.~\ref{eq:lemma-ZO-3} in lemma~\ref{lemma:ZO}, 
    2) applies $\normsq{a+b} \le 2\normsq{a} + 2\normsq{b}$, 
    3) applies assumption~\ref{assum:smoothness} (smoothness), 
    4) applies $\BE(X^2) = \BE(X)^2 + \Var(X)$ and assumption~\ref{assum:bounded_variance} (bounded variance).

%%%%%%%%%%%%%%%%%%%%%%%%%%%%%%%%%%%%%%%%%%%%%%%%%%%%%%%%%%%%%%%%%%%% Combine the gradient %%%%%%%%%%%%%%%%%%%%%%%%%%%%%%%%%%%%%%%%%%%%%%%%%%%%%%%%%%%%%%%%%%%
\subsection{Combine the gradient} \label{Appendix:sec:combine_gradient}
    Start with the Eq.~\ref{eq:global_step}, additionally taking expectation w.r.t. activated client $\mt$, and applying the assumption~\ref{assum:independent_client} (independent client). 

    \begin{align}\label{eq:global_step_combine}
                  & \BE[\mt, i, u] \bigsb{ \f{w_0^{t+1}, w_1^{t}, \cdots,  w_\mt^{t+1}, \cdots, w_M^{t}}  -  \f{w_0^{t}, w_1^{t}, \cdots,  w_\mt^{t}, \cdots, w_M^{t}}} \NN
        \leos{}   & - \bigp{\frac{1}{2} \eta_0 - L \eta_0^2 }\BE[i] \normsq{\nz \f{w_0^{t}, \bw^{t}} } - \summ p_m \bigp{\frac{1}{2} \eta_m - 2 L \eta_m^2 d_m } \BE[i, u] \normsq{\nmt \f{w_0^{t}, \bw^{t}} } \NN
                  & + \bigsb{\bigp{ \frac{1}{2} \eta_0 + L \eta_0^2}L_0^2 + \summ p_m \bigp{\eta_m + 2 L \eta_m^2 d_m}L_m^2 } \BE[i] \normsq{\bw - \tbw}   \NN
                  & + L \eta_0^2\sigma_0^2 + \summ p_m 2 L \eta_m^2 d_m \sigma_m^2 +  \summ p_m \frac{1}{4} \bigp{ L \eta_m^2 + \eta_m } \mu_m^2 L_m^2 d_m^2 \NN
        \leos{1)} & - \bigp{\frac{1}{2} \eta_0 - L \eta_0^2 }\BE[i] \normsq{\nz \f{w_0^{t}, \bw^{t}} } - \summ p_m \bigp{\frac{1}{2} \eta_m - 2 L \eta_m^2 d_m } \BE[i, u] \normsq{\nmt \f{w_0^{t}, \bw^{t}} } \NN
                  & + \bigsb{\bigp{ \frac{1}{2} \eta_0 + L \eta_0^2}L_0^2 + \summ p_m \bigp{\eta_m + 2 L \eta_m^2 d_m}L_m^2 } \BE[i] \normsq{\bw - \tbw}   \NN
                  & + Q_1 \NN
        \leos{2)} & - \frac{1}{4} \eta_0 \BE[i] \normsq{\nz \f{w_0^{t}, \bw^{t}} } - \summ p_m \frac{1}{4} \eta_m \BE[i, u] \normsq{\nmt \f{w_0^{t}, \bw^{t}} } \NN
                  & + \bigsb{\bigp{ \frac{1}{2} \eta_0 + L \eta_0^2}L_0^2 + \summ p_m \bigp{\eta_m + 2 L \eta_m^2 d_m}L_m^2 } \BE[i] \normsq{\bw - \tbw}  + Q_1 \NN
        \leos{3)} & - \frac{1}{4} \min \bigcb{\eta_0, p_m \eta_m} \BE[i] \normsq{\nabla \f{w_0^{t}, \bw^{t}} }  \NN
                  & + \bigsb{\bigp{ \frac{1}{2} \eta_0 + L \eta_0^2}L_0^2 + \summ p_m \bigp{\eta_m + 2 L \eta_m^2 d_m}L_m^2 } \BE[i] \normsq{\bw - \tbw}  + Q_1 \NN
    \end{align} 
    where 1) to simplify the notation, define $Q_1$ to substitute the last row,
    2) let $\eta_0 \le \frac{1}{4L}$ then $ - \frac{1}{2} \eta_0 + L \eta_0^2 < - \frac{1}{4} \eta_0 $, and let $ \eta_m \le \frac{1}{4 L d_m}$, then $ \frac{1}{2} \eta_0 - L \eta_0^2 \le \frac{1}{4} \eta_0$ and $\frac{1}{2} \eta_m - 2 L \eta_m^2 d_m \le \frac{1}{4} \eta_m$, 
    3) uses the orthogonality of $\nabla f$, i.e. $  \normsq{ \nabla \f{ w_0, \bw}} = \normsq{\nz\f{w_0, \bw}} + \sum_{m=1}^M \normsq{ \nm \f{w_0, \bw} }$. 

%%%%%%%%%%%%%%%%%%%%%%%%%%%%%%%%%%%%%%%%%%%%%%%%%%%%%%%%%%%%%%%%%%%% Lyapunov Function %%%%%%%%%%%%%%%%%%%%%%%%%%%%%%%%%%%%%%%%%%%%%%%%%%%%%%%%%%%%%%%%%%%
\subsection{Define the Lyapunov function to eliminate the client's delay. } \label{Appendix:subsec:Lyapunov}
    Define a Lyapunov function. 
    \begin{align} \label{eq:Lyapunov}
        M^t = \f{w_0^t, \bw^t} + \sumitau \theta_i \norm{\bw^{\tpo - i} - \bw^{t-i} }^2
    \end{align} 

    Taking expectation w.r.t. the activated client $\mt$, sample index $i$, and the random direction $u$. 
    \begin{align} \label{eq:Differ_Lyapunov}
                        & \BE \bigp{M^\tpo - M^t} \NN
            \eq         & \BE \bigsb{\f{w_0^\tpo, \bw^\tpo} + \sumitau \theta_i \normsq{\bw^{\tpo +1 - i}-\bw^{t+1-i} }}  - \BE\bigsb{\f{w_0^t, \bw^t} + \sumitau \theta_i \normsq{\bw^{\tpo - i} - \bw^{t-i} }} \NN
            \eq         & \BE \bigsb{\f{w_0^\tpo, \bw^\tpo} - \f{w_0^t, \bw^t} }  + \sumitau \theta_i \BE \normsq{\bw^{\tpo +1 - i}-\bw^{t+1-i} } - \sumitau \theta_i \normsq{\bw^{\tpo - i} - \bw^{t-i} }\NN
            \leos{1)}   & - \frac{1}{4} \min\bigcb{\eta_0, p_m \eta_m} \BE \normsq{\nabla \f{w_0^{t}, \bw^{t}} }  + Q_1 \NN
                        & + \bigsb{\bigp{ \frac{1}{2} \eta_0 + L \eta_0^2}L_0^2 + \summ p_m \bigp{\eta_m + 2 L \eta_m^2 d_m}L_m^2 } \uba{\BE \normsq{\tbw^{t} - \bw^{t}}}  \NN
                        & + \ubb{\sumitau \theta_i \BE \normsq{\bw^{\tpo +1 - i}-\bw^{t+1-i} } - \sumitau \theta_i \normsq{\bw^{\tpo - i} - \bw^{t-i} }}\NN
            \leos{2)}   & - \frac{1}{4} \min\bigcb{\eta_0, p_m \eta_m} \BE \normsq{\nabla \f{w_0^{t}, \bw^{t}} }  + Q_1 \NN
                        & + \bigsb{\bigp{ \frac{1}{2} \eta_0 + L \eta_0^2}L_0^2 + \summ p_m \bigp{\eta_m + 2 L \eta_m^2 d_m}L_m^2 } \tau \sumitau \BE\normsq{ \bw^{t+1-i} - \bw^{t-i}}  \NN
                        & + \theta_1 \BE \norm{\bw^\tpo - \bw^t}^2 + \sumitau[-1] (\theta_\ipo - \theta_i) \BE \norm{\bw^{t +1 - i} - \bw^{t-i} }^2 - \theta_\tau \BE \norm{\bw^{\tpo - \tau} - \bw^{t-\tau} }^2 \NN
            \le         & - \frac{1}{4} \min\bigcb{\eta_0, p_m \eta_m} \BE \normsq{\nabla \f{w_0^{t}, \bw^{t}} }   + Q_1 \NN
                        & + \theta_1 \BE \norm{\bw^\tpo - \bw^t}^2 \NN
                        & + \sumitau[-1] \bigp{\theta_\ipo - \theta_i + \bigsb{\bigp{ \frac{1}{2} \eta_0 + L \eta_0^2}L_0^2 + \summ p_m \bigp{\eta_m + 2 L \eta_m^2 d_m}L_m^2 } \tau} \BE \normsq{\bw^{t +1 - i} - \bw^{t-i} } \NN
                        & - \bigcb{ \theta_\tau - \bigsb{\bigp{ \frac{1}{2} \eta_0 + L \eta_0^2}L_0^2 + \summ p_m \bigp{\eta_m + 2 L \eta_m^2 d_m}L_m^2 } \tau}\BE \normsq{\bw^{\tpo - \tau} - \bw^{t-\tau} } \NN
        \end{align}

    where 1) plugging in Eq.~\ref{eq:global_step_combine},  2) plugging in a) and b). 

    For a) in Eq.~\ref{eq:Differ_Lyapunov}:
    \begin{align} 
         \BE\normsq{\tbw^t - \bw^t} \leos{1)} \BE\normsq{ \sumitau \bigp{\bw^\ipo - \bw^i }} \leos{2)} \tau \sumitau \BE\normsq{ \bw^{t+1-i} - \bw^{t-i} }
    \end{align}
    where 1) applies assumption~\ref{assum:bound_delay} (uniformly bounded delay), 2) applies Cauchy-Schwarz inequality, i.e. $ \bigp{\sum_{i=0}^{n-1} x_i }^2 = \bigp{\sum_{i=0}^{n-1} 1 \cdot x_i}^2 \le n \sum_{i=0}^{n-1} x_i^2$. 

    For b) in Eq.~\ref{eq:Differ_Lyapunov}:
    
    \begin{align}
            & \sumitau \theta_i \BE \norm{\bw^{\tpo +1 - i}-\bw^{t+1-i} }^2 - \sumitau \theta_i \BE \norm{\bw^{\tpo - i} - \bw^{t-i} }^2 \NN
        \eq & \theta_1 \BE \norm{\bw^\tpo - \bw^t}^2 + \sumitau[-1] (\theta_\ipo - \theta_i) \BE \norm{\bw^{t +1 - i} - \bw^{t-i} }^2 - \theta_\tau \BE \norm{\bw^{\tpo - \tau} - \bw^{t-\tau} }^2
    \end{align}

    Let $ \theta_1 = \tau^2 \bigsb{\bigp{ \frac{1}{2} \eta_0 + L \eta_0^2}L_0^2 + \summ p_m \bigp{\eta_m + 2 L \eta_m^2 d_m}L_m^2 } $ and define the recursive formula for $\theta_i$:
    \begin{align}
        \theta_\ipo = \theta_i - \tau Q_1 
    \end{align}
    if follows that: 

    \begin{align}
               &\theta_\tau - \tau \bigsb{\bigp{ \frac{1}{2} \eta_0 + L \eta_0^2}L_0^2 + \summ p_m \bigp{\eta_m + 2 L \eta_m^2 d_m}L_m^2 } \NN
             = & \theta_1 - \tau^2 \bigsb{\bigp{ \frac{1}{2} \eta_0 + L \eta_0^2}L_0^2 + \summ p_m \bigp{\eta_m + 2 L \eta_m^2 d_m}L_m^2 } = 0 \NN
    \end{align}

    Then Eq.~\ref{eq:Differ_Lyapunov} becomes
    
    \begin{align}\label{eq:Differ_Lyapunov-1}
                  & \BE \bigp{M^\tpo - M^t} \NN
        \le       & - \frac{1}{4} \min\bigcb{\eta_0, p_m \eta_m} \BE \normsq{\nabla \f{w_0^{t}, \bw^{t}} }   + Q_1 \NN
                  & + \tau^2 \bigsb{\bigp{ \frac{1}{2} \eta_0 + L \eta_0^2}L_0^2 + \summ p_m \bigp{\eta_m + 2 L \eta_m^2 d_m}L_m^2 } \ubc{\BE \norm{\bw^\tpo - \bw^t}^2} \NN
        \leos{1)} & - \frac{1}{4} \min\bigcb{\eta_0, p_m \eta_m} \BE \normsq{\nabla \f{w_0^{t}, \bw^{t}} }   + Q_1 \NN
                  & + \tau^2 \bigsb{\bigp{ \frac{1}{2} \eta_0 + L \eta_0^2}L_0^2 + \summ p_m \bigp{\eta_m + 2 L \eta_m^2 d_m}L_m^2 } \summ p_m \eta_m^2 \bigp{ 2d_m \mG_m^2 + \frac{1}{2} \mu_m^2 L_m^2 d_m^2 } \NN
        \leos{2)} & - \frac{1}{4} \min\bigcb{\eta_0, p_m \eta_m} \BE \normsq{\nabla \f{w_0^{t}, \bw^{t}} }   + Q_1 \NN
                  & + Q_2 \NN
    \end{align}
    where 1) plugs in c), 
    2) simplify the notation by denoting the second line as $Q_2$. 
    
    For c): 
    \begin{align}
                  & \BE[\mt, i, u] \normsq{\bw^\tpo - \bw^t} \NN
        \eqos{1)} & \BE[\mt, i, u] \eta_\mt^2 \normsq{ \hn_\mt \f[i]{w_0^{t}, \tbw^t}  } \NN
        \leos{2)} & \BE[\mt, i, u] \eta_\mt^2 \bigp{ 2d_\mt \normsq{ \nmt \f[i]{w_0^{t}, \tbw^t}} + \frac{1}{2} \mu_\mt^2 L_\mt^2 d_\mt^2 } \NN
        \leos{3)} & \BE[\mt, i, u] \eta_\mt^2 \bigp{ 2d_\mt \mG_\mt^2 + \frac{1}{2} \mu_\mt^2 L_\mt^2 d_\mt^2 } \NN
        \leos{4)} & \summ p_m \eta_m^2 \bigp{ 2d_m \mG_m^2 + \frac{1}{2} \mu_m^2 L_m^2 d_m^2 } \NN
    \end{align}
    
    where 1) the update rule for the communication round, 
    2) applies Eq.~\ref{eq:lemma-ZO-3} in lemma~\ref{lemma:ZO}, 
    3) applies assumption~\ref{assum:bounded_embedding_gradients} (bounded block-coordinated gradient), 
    4) applies assumption~\ref{assum:independent_client} (independent client).

%%%%%%%%%%%%%%%%%%%%%%%%%%%%%%%%%%%%%%%%%%% 
\subsection{Bound the gradient $\nabla \f{w_0^t, \bw^t}$} \label{Appendix:subsec:Bound_the_gradient}
    Start with Eq.~\ref{eq:Differ_Lyapunov-1}:
    \begin{align} 
                    & \BE \bigp{M^\tpo - M^t} \NN
        \le         & - \frac{1}{4} \min\bigcb{\eta_0, p_m \eta_m} \BE \normsq{\nabla \f{w_0^{t}, \bw^{t}} }   + Q_1 + Q_2     
    \end{align}

    Summing over the global iteration $t = 0, 1, ...T-1$, arrange the equation and divided it by $T$ from both sides. 
    \begin{align} 
                    &  \frac{1}{4T} \min\bigcb{\eta_0, p_m \eta_m} \sumt \BE \normsq{\nabla \f{w_0^{t}, \bw^{t}} }   \NN
        \le         &   \frac{\BE\bigp{M^0 - M^T}}{T} + Q_1 + Q_2 \NN  
        \leos{1)}   &   \frac{\BE\bigp{f^0 - f^*}}{T} + Q_1 + Q_2\NN       
    \end{align}
    where 1) applies $\BE \bigp{M^0 - M^{T}} = \f{w_0^0, \bw^0} - \f{w_0^T, \bw^T} - \sumitau \theta_i \norm{\bw^{T - i} - \bw^{T-i} }^2 \le \f{w_0^0, \bw^0} - \f{w_0^T, \bw^T} \le  f^0 - f^* $, we use $f^0$ to denote $\f{w_0^0, \bw^0}$ and applying assumption~\ref{assum:feasible_optimal_solution}. 
    % if the assumption about the optimal is added here. 
    
    Dividing $ \zeta = \frac{1}{4} \min\bigcb{\eta_0, p_m \eta_m}  $ from both sides: 
    \begin{align} \label{eq:result-1}
                    & \frac{1}{T } \sumt \BE \normsq{\nabla \f{w_0^{t}, \bw^{t}} }   \NN          
        \le         & \frac{\BE\bigp{f^0 - f^*}}{T\zeta} + \frac{ Q_1}{\zeta} + \frac{Q_2}{\zeta } \NN
        \leos{1)}   & \frac{\BE\bigp{f^0 - f^*}}{T\zeta} \NN
                    & + \frac{1}{\zeta} L \eta_0^2\sigma_0^2 + \frac{1}{\zeta} \summ p_m 2 L \eta_m^2 d_m \sigma_m^2 + \frac{1}{\zeta} \summ p_m \frac{1}{4} \bigp{ L \eta_m^2 + \eta_m } \mu_m^2 L_m^2 d_m^2  \NN 
                    & + \frac{1}{\zeta} \tau^2 \bigsb{\bigp{ \frac{1}{2} \eta_0 + L \eta_0^2}L_0^2 + \summ p_m \bigp{\eta_m + 2 L \eta_m^2 d_m}L_m^2 } \summ p_m \eta_m^2 \bigp{ 2d_m \mG_m^2 + \frac{1}{2} \mu_m^2 L_m^2 d_m^2 } \NN 
    \end{align}
    where 1) plugs in a) and b). 

    To simplify the result, let  $\Ls = \max_m \bigcb{L, L_0, L_m}$, $ \ds = \max_m \bigcb{ d_{m} } $, $\eta_0 = \eta_m = \eta \le \frac{1}{4 \Ls \ds}$, $\frac{1}{\ps} = \min_m p_m $, $\mus = \max_m \bigcb{\mu_m} $, $\mGs = \max_m \bigcb{\mG_m}$, then $ \zeta = \frac{1}{4} \min\bigcb{\eta_0, p_m \eta_m} = \frac{\eta}{4 \ps}$. Eq.~\ref{eq:result-1} can be further simplified:
    \begin{align} \label{eq:result-2}
                    & \frac{1}{T } \sumt \BE \normsq{\nabla \f{w_0^{t}, \bw^{t}} }  \NN
        \leos{1)}   & \frac{ 4\ps \BE \bigp{f^0 - f^*}}{T \eta} \NN
                    & + \frac{4 \ps}{\eta} L \eta_0^2\sigma_0^2 + \frac{4 \ps}{\eta} \summ p_m 2 L \eta_m^2 d_m \sigma_m^2 + \frac{4 \ps}{\eta} \summ p_m \frac{1}{4} \bigp{ L \eta_m^2 + \eta_m } \mu_m^2 L_m^2 d_m^2  \NN 
                    & + \frac{4 \ps}{\eta} \tau^2 \bigsb{\bigp{ \frac{1}{2} \eta_0 + L \eta_0^2}L_0^2 + \summ p_m \bigp{\eta_m + 2 L \eta_m^2 d_m}L_m^2 } \summ p_m \eta_m^2 \bigp{ 2d_m \mG_m^2 + \frac{1}{2} \mu_m^2 L_m^2 d_m^2 } \NN 
        \leos{2)}   & \frac{ 4\ps \BE \bigp{f^0 - f^*}}{T \eta} \NN
                    & + \frac{4 \ps}{\eta} L \eta_0^2\sigma_0^2 + \frac{4 \ps}{\eta} \summ p_m 2 L \eta_m^2 d_m \sigma_m^2 + \frac{4 \ps}{\eta} \summ p_m \frac{1}{4} \bigp{ L \eta_m^2 + \eta_m } \mu_m^2 L_m^2 d_m^2  \NN 
                    & + \frac{4 \ps}{\eta} \tau^2 \bigsb{\bigp{ \frac{1}{2} \eta_0 + \frac{1}{4} \eta_0}L_0^2 + \summ p_m \bigp{\eta_m + \frac{1}{2} \eta_m}L_m^2 } \summ p_m \eta_m^2 \bigp{ 2d_m \mG_m^2 + \frac{1}{2} \mu_m^2 L_m^2 d_m^2 } \NN 
        \leos{3)}   & \frac{ 4\ps \BE \bigp{f^0 - f^*}}{T \eta} \NN
                    & + {4 \ps} \Ls \eta \sigma_*^2 + 8 \ps \Ls \eta \ds \sigma_*^2 + \ps \Ls \eta \mus^2 \Ls^2 \ds^2 + \ps \Ls \mus^2 \Ls^2 \ds^2 \NN 
                    & + \ps \tau^2 \bigp{\frac{9}{2} \eta} \Ls^2 \eta \bigp{ 4 \ds \mGs^2 + \mus^2 \Ls^2 \ds^2 } \NN 
        \leos{4)}   & \frac{4\ps \BE \bigp{f^0 - f^*}}{T \eta} \NN
                    & + \eta \bigp{ 4 \ps \Ls \sigma_*^2 + 8 \ps \Ls \ds \sigma_*^2 + \ps \Ls \mus^2 \Ls^2 \ds^2} \NN
                    & + \eta^2\bigp{ \frac{9}{2} \ps \tau^2  \Ls^2 \bigp{ 4 \ds \mGs^2 + \mus^2 \Ls^2 \ds^2 } }\NN 
                    & + \mus^2\bigp{\ps \Ls  \Ls^2 \ds^2} \NN 
        \leos{5)}   & \frac{4\ps \BE \bigp{f^0 - f^*}}{T \eta} \NN 
                    & + \eta \bigp{ 4 \ps \Ls \sigma_*^2 + 8 \ps \Ls \ds \sigma_*^2 + \ps \Ls \mus^2 \Ls^2 \ds^2} \NN
                    & + \eta^2\bigp{ 18 \ps \tau^2  \Ls^2 \ds \mGs^2 + 5 \ps \tau^2 \Ls^2 \mus^2 \Ls^2 \ds^2 }\NN 
                    & + \mus^2\bigp{\ps \Ls^3 \ds^2} \NN
    \end{align}

    Where 1) plugs in the above variables for $ \zeta$,
    2) simplify by $\eta_0 \le \frac{1}{4L} $ and $\eta_m \le \frac{1}{4L d_m} $
    3) plugs in the variables $ \eta, \mus, \Ls $,
    4) collect the $\eta$ and $\mu$, 
    5) simply $\frac{9}{2} < 5$. 

    The proof of Theorem \ref{theo:1} is complete. \hfill $\blacksquare$ 

    Suppose we set $ \eta = \frac{1}{\sqrt{T}} $, and $\mu = \frac{1}{\sqrt{T}}$, the above equation becomes:

    \begin{align}
                    &   \frac{1}{T} \sumt \BE \normsq{\nabla \f{w_0^{t}, \bw^{t}} }  \NN
        \leos{5)}   &   \frac{1}{\sqrt{T}} \bigsb{ 4\ps \BE \bigp{f^0 - f^*} +  4 \ps \Ls \sigma_*^2 + 8 \ps \Ls \ds \sigma_*^2 } \NN
                    & + \frac{1}{T}\bigp{ 18 \ps \tau^2  \Ls^2 \ds \mGs^2 + 5 \ps \tau^2 \mus^2 \Ls^4 \ds^2 + \ps  \Ls^3 \ds^2} \NN 
                    & + \frac{1}{T^\frac{3}{2}} \bigp{ \ps \Ls^3 \ds^2 } \NN
    \end{align}

    Therefore, 
    \begin{align}
        \frac{1}{T} \sumt \BE \normsq{\nabla \f{w_0^{t}, \bw^{t}} } = \mathcal{O}\bigp{\frac{d}{\sqrt{T}}}
    \end{align}
    where $d = d_* = \underset{m}{\max} \bigcb{d_{m}}$ (for clear notation), $T$ is the number of iterations.

    %%% End. 
    The proof of Corollary \ref{coro:1} is complete. \hfill $\blacksquare$

\section{Extra Experiment Details}
    The algorithm for Syn-ZOO-VFL:
    \begin{algorithm}[H] 
                \caption{The Synchronous Modification of ZOO-VFL~\cite{zhang2021desirable}}
                \label{algo:syn-ZOO-VFL}
                \begin{algorithmic}[1]
                    \item Initialize variables for workers $m \in [M]$
                    \FOR {$t = 0, ..., T-1 $}
                        \STATE Random sample a sample $i$ (or batch $B$).
                        \FOR {client $m$ in $[M]$ in parallel} 
                            \STATE Client $m$ compute and send $ h_{m, i} = h_m(w_m; x_{m, i})$ and $ \hat{h}_{m, i} = h_m(w_m+\mu \bu_{m, i}; x_{m, i})$ to the server. 
                            \STATE The server calculates $ \delta_m = f_i(w_0, ... \hat{h}_{m, i}...) - f_i(w_0, h_{1, i}, ... h_{M, i}) $ and send back to the client. 
                            \STATE Client $m$ calculate the stochastic gradient w.r.t. its local parameter $w_m$ with the $\delta_m$ received from the server: $ \hn_\wm \f[i]{\cdot} = \frac{\phi(d_m)}{\mu} \delta_m \bu_{m, i} $
                            \STATE Client $m$ update its parameter with gradient descent $w_m \leftarrow w_m - \eta_m \hn_\wm \f[i]{\cdot}$
                        \ENDFOR
                        \STATE The server calculates its local stochastic gradient estimation via $ \hn_{w_0}\f[i]{\cdot} = \frac{\phi(d_0)}{\mu}\bigsb{ f_i(w_0+\mu \bu_{0, i}, ... \hat{h}_{m, i}...) - f_i(w_0, h_{1, i}, ... h_{M, i} ;y_i) } \bu_{0, i} $
                        \STATE The server update its local parameter with gradient descent $ w_0 \leftarrow w_0 - \eta_0 \hn_{w_0}\f[i]{\cdot} $
                    \ENDFOR
                \end{algorithmic}
            \end{algorithm}

\end{document}